\documentclass[letterpaper, 10 pt, conference]{ieeeconf}  

\IEEEoverridecommandlockouts                              

\usepackage{cite}
\usepackage{amsmath,amssymb,amsfonts}
\usepackage{algorithm}
\usepackage{float}
\usepackage{algorithmicx}
\usepackage{algpseudocode}
\usepackage{graphicx}
\usepackage{textcomp}
\usepackage{subfig}

\title{\LARGE \bf
Model-based Decision Making with Imagination\\
for Autonomous Parking*
}

\author{Ziyue Feng$^{1}$, Shitao Chen$^{2}$, Yu Chen$^{3}$ and Nanning Zheng$^{4}$, Fellow, IEEE
\thanks{*This research was partially supported by the National Natural Science Foundation of China (No. 61773312, 61790563), the Programme of Introducing Talents of Discipline to University (No. B13043).}
\thanks{Ziyue Feng$^{1}$, Shitao Chen$^{2}$, Yu Chen$^{3}$ are with Institute of Artificial Intelligence and Robotics in Xi'an Jiaotong University, Xi'an, Shannxi, P.R.China
	{\tt\small e-mail:\{brother\_yue, chenshitao, alan19960212\}@stu.xjtu.edu.cn}}%
\thanks{Nanning Zheng$^{4}$ is the director of Institute of Artificial Intelligence and Robotics, Xi'an Jiaotong University, Xi'an, Shannxi, P.R.China
	{\tt\small Correspondence: nnzheng@mail.xjtu.edu.cn}}%
}

\begin{document}

\captionsetup[figure]{labelfont={bf},labelformat={default},labelsep=period,name={Fig.}}

\maketitle
\thispagestyle{empty}
\pagestyle{empty}

\begin{abstract}

Autonomous parking technology is a key concept within autonomous driving research. This paper will propose an imaginative autonomous parking algorithm to solve issues concerned with parking. The proposed algorithm consists of three parts: an imaginative model for anticipating results before parking, an improved rapid-exploring random tree (RRT) for planning a feasible trajectory from a given start point to a parking lot, and a path smoothing module for optimizing the efficiency of parking tasks. Our algorithm is based on a real kinematic vehicle model; which makes it more suitable for algorithm application on real autonomous cars. Furthermore, due to the introduction of the imagination mechanism, the processing speed of our algorithm is ten times faster than that of traditional methods, permitting the realization of real-time planning simultaneously. In order to evaluate the algorithm's effectiveness, we have compared our algorithm with traditional RRT ,within three different parking scenarios. Ultimately, results show that our algorithm is more stable than traditional RRT and performs better in terms of efficiency and quality.

\end{abstract}

\begin{figure*}
	\centering
	\includegraphics[width=1\textwidth]{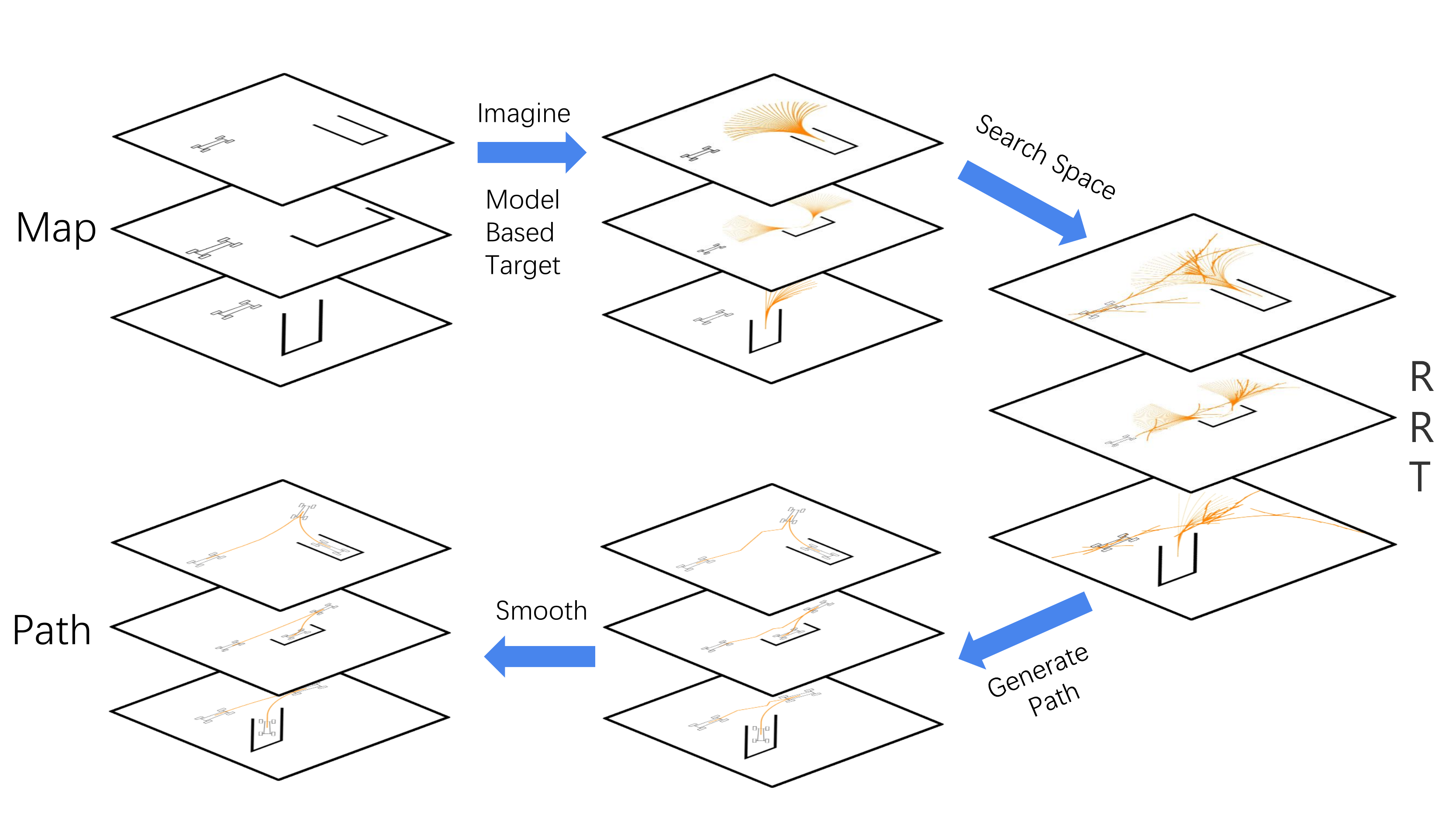}
	\caption{The main workflow of our algorithm. The parking task has three scenarios: perpendicular, parallel and echelon. Imagine the car is already parked in the parking lot, use the algorithm that corresponds to the specific parking scenario to build a target tree, and then use the modified RRT to search the configure space and connect paths from RRT tree and target tree. After that, the trajectory is generated. Once that is completed we use our smoothing algorithm to smooth the generated trajectory under vehicle kinematic constraints.}
	\label{fig:method_big}
\end{figure*}

\section{INTRODUCTION}

Self-parking is one of the key technologies to achieve autonomous driving. The Society of Automotive Engineers (SAE) J3016 standard defines five distinct levels of vehicle automation, including the parking assist technologies. The first three are progressions of assisted parking, i.e., driver assistance, partial automation and conditional automation. The fourth level is a near-automated technology called high automation; while the last one is totally automated, which is defined as full automation. Currently, the parking capabilities of a number of vehicles on the road nowadays are among level 1 and level 2. It is commonly recognized that autonomous parking is logically the first step towards commercial applications of self-driving, with fully automated vehicles. At present we are not yet at a stage where autonomous parking is completely achievable. However, many researchers are attempting to reach this stage through continuous research. Most car manufacturers and technology companies in the automation industry are expecting fully automated vehicles to be available by 2020, with self-parking as a primary feature.

For the realization of self-parking, the parking-planning task is that, when given a parking lot, we need to find the optimal collision free trajectory, under vehicle kinematic constraints to maneuver a car into a target place. Yet, there are several difficulties involved in this task. The combination of both moving forward actions and backward in parking planning means this process is more complex than on-road planning. Sometimes moving backwards slightly can make it easier for the car to maneuver to a narrower place. However, a negative result of this action is that it could lead to a decreased human-like trajectory performance, as well as a uncomfortable passenger experience. Moreover, the turning angle of the parking trajectory is typically bigger than on-road trajectory, therefore  the corridor is much more narrow, which makes the influence caused by the difference of radius between inner wheels greater when moving forward, this  is the same for the outer wheels, when moving backward.

The RRT algorithm is a popular method to solve motion planning problems. It was first proposed by Steven \emph{et al.} \cite{RRT}, and has then been applied to the parking-planning task \cite{2011RRT}. This method constructs a tree to store some accessible places from the start point, and then it randomly grows the tree to explore the whole space until the target point is close enough to the tree. Finally, a feasible trajectory will be generated. The RRT can easily take into account and handle complex environments with obstacles.

However, there is still a long way to go between the RRT and the practical parking planning tasks. When the car gets closer to the target point, a little bit of bias reduction needs to be executed with quite a few steps. This causes the car getting close to the garage quickly but then moves forward and backward over and over again in order to adjust to a perfect position. The whole parking process with RRT takes a long time and makes the final part of the trajectory complex. Bi-RRT \cite{MISC} uses tree growing from target to reduce the search time, but the final part of the trajectory is still complex. We imagine that the car is already parked in the parking lot and then we try to drive the car out of the parking lot with various pre-defined driving strategies. After the operation we will obtain dozens of feasible driving-out paths, each with twenty nodes. By reversing these driving-out paths to parking-in paths, there will be a tree with dozens of paths and hundreds of nodes. Then we expand the target point, in RRT algorithm, to this well-defined model-based target tree. Once RRT tree reaches any node of the target tree, the final trajectory will be generated. Since the model-based target tree is well defined it results in the trajectory in target tree being very smooth. The car can get into the target position in one movement. With this target expansion, the RRT tree can get close to the target, with fewer steps and less time cost. In fact, model-based planning is more than ten times quicker than traditional RRT. Our algorithm is not only better but is more stable than traditional RRT, in planning time and quality. RRT still has another drawback, that its trajectory can be extremely complex. This is because every edge of the tree is generated towards a random direction. There are some methods to smooth the trajectory, but they do not take into account the kinematic constraints of a real car. This may not be a problem with on-road planning, but since the parking lot is much narrower than an on-road environment, the smoothed parking trajectory may be unfeasible. We propose an algorithm to smooth the trajectory obeying the kinematic constraints. The trajectory smoothed via our algorithm has a guaranteed feasibility. With these improvements, our algorithm can be applied in actual parking planning task.

\section{Related Work}
Path planning for parking tasks \cite{inproceedings} means to find a collision-free path from a given start point to a final target point in a parking lot. The very beginning stages of developing path planning methods is the car of Dubins \emph{et al.} \cite{dubin}, as well as that of Reeds and Shepp \emph{et al.} \cite{reeds} in the domain of robotics. However, these two methods do not consider obstacles and continuous turning angles. Then some research have been done in the field of motion planning. Methods of motion planning for automated vehicles can be divided into four groups \cite{motion_planing_review}, including graph search-based planners, for example, Dijkstra algorithm \cite{dij1}, \cite{dij2} and A-Star algorithm \cite{astar}, sampling-based planners like RRT \cite{RRT}, deep learning-based planners like \cite{dl}, \cite{chen2017brain}, \cite{chen2017cognitive} and numerical optimization like \cite{numopt}. Our method is based on RRT, which belongs to the second group.
RRT is one of the sampling-based planning methods. It has many improved branches, like Bi-RRT \cite{MISC}, RRT* \cite{RRTstar}, CL-RRT \cite{clRRT} and DD-RRT \cite{ddRRT}. It has also been applied to parking planning task in \cite{2011RRT}. Our method has achieved many improvements to the basic RRT. We expand the target point to a model-based target tree. This expansion allows for the generated path to be more smooth and human-like. However, a downside to the sampling-based approach is the uncertainty of planning time. Our expansion can make the planing time shorter and more stable.
Another drawback of every sample-based planning algorithm is that every segment of the path is generated with random direction and target, which which results in the path being more complex and with some redundant nodes. Some smoothing algorithms already exist for this particular problem, including, cubic polynomials \cite{cubic}, quintic polynomials \cite{quintic}, \cite{quintic2}, Bezier curves \cite{bezier}, \cite{bezier3}, \cite{bezier4}, B-splines \cite{bsp} and Clothoids \cite{clothiods}. All of these smoothing algorithms have a common drawback: they violently distort the path into a smooth curve without considering the kinematic constraints of a real car. In this paper, we propose a new smoothing algorithm that takes the vehicle kinematic constraints into consideration. The path smoothed by our algorithm is guaranteed to be practical.

\section{Method}
We divide the parking tasks into three scenarios: perpendicular parking, parallel parking and echelon parking. Imagine the car is already parked in the parking lot, we define three different model-based target trees with three different scenarios. Each target tree has dozens of paths and hundreds of nodes. We then go on to expand the target point in RRT algorithm to this tree. Once the RRT tree reaches any node of the target tree the final trajectory is generated. This trajectory consists of two parts: a part from RRT tree and a part from the target tree. The model-based target tree makes the latter extremely smooth and enables a processing time of our algorithm more than ten times faster than the traditional RRT. Our algorithm is not only more efficient but more stable in planning time and quality. We use our smoothing algorithm to smooth the former part of the trajectory under vehicle kinematic constraints. Different from other smoothing algorithms, trajectories smoothed by our algorithm have guaranteed feasibility. This procedure is illustrated in Fig. \ref{fig:method_big}.

\subsection{Point Pursuit}

\begin{figure}[tb]
	\centering
	\includegraphics[width=0.45\textwidth]{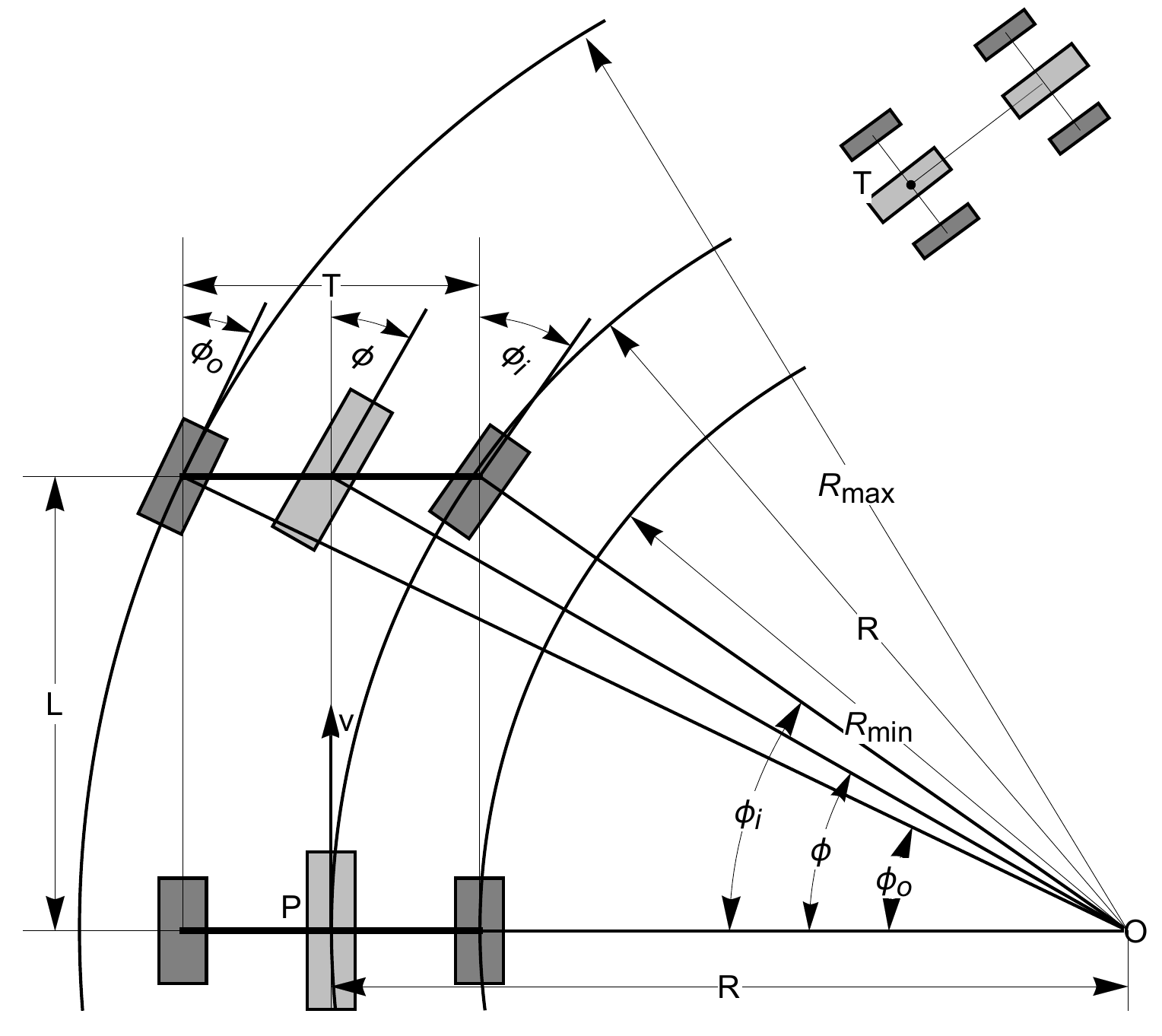}
	\caption{Kinematic model in `point pursuit' algorithm. It is based on Ackerman model. A car has four wheels and two equivalent wheels. The rear equivalent wheel determines the car’s coordinates. In this figure, the car is positioned at the start point $(0,0,\frac{\pi}{2})$.}
	\label{fig:ackerman}
\end{figure}

In this paper, a basic algorithm is called `point pursuit', which calculates the best turning angle to maneuver the car to get it closer to a particular point. Consider a vector $(X,Y,\theta)$ to indicate a car's state. X and Y is its position in a 2D surface, $\theta$ is the orientation. As shown in Fig. \ref{fig:ackerman}, the car's starting point is $ P(0,0,\frac{\pi}{2})$, the target point is $T(X_t,Y_t,\theta_t)$. The car will move 0.1 meters with a fixed turning angle $\phi$ to point $A(X_a,Y_a,\theta_a)$. Therefore, we need to find the best $\phi$ to make the distance between P and A as small as possible. The $\phi$ mentioned previously is an equivalent turning angle calculated by outer wheel turning angles $\phi_o$ and inner wheel turning angle $\phi_i$.

\begin{equation}
X_a=R(1-\cos\frac{Vt}{R})
\end{equation}
\begin{equation}
Y_a=R \sin \frac{Vt}{R}
\end{equation}
\begin{equation}
\theta_a = \frac{\pi}{2}-\frac{Vt}{R}
\end{equation}
\begin{equation}
R=\frac{L}{\tan \phi}
\end{equation}
\begin{equation}
V=0.1m/s,\ t=1s
\end{equation}
\begin{equation}
distance=(X_a-0)^{2}+(Y_a-0)^{2}+(\theta_a - \frac{\pi}{2})^{2}
\end{equation}

Algorithm `point pursuit' will compute the best $\phi$ to minimize the distance. If this movement isn't collision-free, we will decrease $\phi$ to find a smaller collision-free $\phi_s$ and increase $\phi$ to find a larger collision-free $\phi_l$. We will choose a better $\phi$ between the two $\phi$ which makes the point A closer to point P. This algorithm is shown in Algorithm \ref{alg1}.

\begin{algorithm}
	\caption{Point Pursuit}
	\label{alg1}
	\begin{algorithmic}
		\Function{Point pursuit}{$N, P$}
		\State $movement\gets argmin\_distance$($N,P$)$ $
		\State $simulate\_movement$($movement$)$ $
		\If{$collide$}
		\State $smaller\gets movement.turning\_angle$
		\State $larger\gets movement.turning\_angle$
		\While{$collide$}
		\State $smaller\gets smaller-little$
		\EndWhile
		\While{$collide$}
		\State $larger\gets larger+little$
		\EndWhile
		\State $ds\gets distance$($moving$($smaller$)$,P$)$ $
		\State $dl\gets distance$($moving$($larger$)$,P$)$ $
		\If{$ds<dl$}
		\State $movement.turning\_angle \gets smaller $
		\Else
		\State $movement.turning\_angle \gets larger $
		\EndIf
		\EndIf
		\State $A\gets movement.destination$
		\State $E \gets movement.path$
		\State \Return $A,\  E$
		\EndFunction
	\end{algorithmic}
\end{algorithm}

\subsection{Basic RRT}

RRT was developed by Steven \emph{et al.} \cite{RRT}, it is an algorithm designed to solve complex high dimensional motion planning problems by randomly building a space-filling tree. Firstly, the start point is the only point of the tree, the growing step is, we randomly sample a point P in free space and then choose the nearest point N in the tree to the sampled point P. Afterwards we find a feasible movement from point N to point A, making A and P be as close as possible. Then we add point A and the movement to the tree as a node and an edge. Repeating this process multiple times, the target point should be close enough to the tree. Then by backtracking the tree, we find a feasible path from the starting point to the target point. This algorithm is presented in Algorithm \ref{alg2}.

\begin{algorithm}
	\caption{Basic RRT}
	\label{alg2}
	\begin{algorithmic}
		\State $Tree\ T \gets startPoint $
		\While{$distance $($ T $,$ target\_point $)$ \ > \  tolerance\_value$}
		\State $P \ \gets \  random\_choice$($ \  $)$ $
		\State $N \ \gets \  nearest\_in\_tree$
		\State $E $,$ \  A \  \gets \  best\_movement $($ N $,$P$)$ $
		\State $T$.$add \  $($ E $,$ A $)$ $
		\EndWhile
		\State $T $.$ backtrack$($\ $)$ $
	\end{algorithmic}
\end{algorithm}

\subsection{Model-based Target}

\begin{figure}[tb]
	\centering
	\subfloat[Perpendicular]{
		\includegraphics[width=0.22\textwidth]{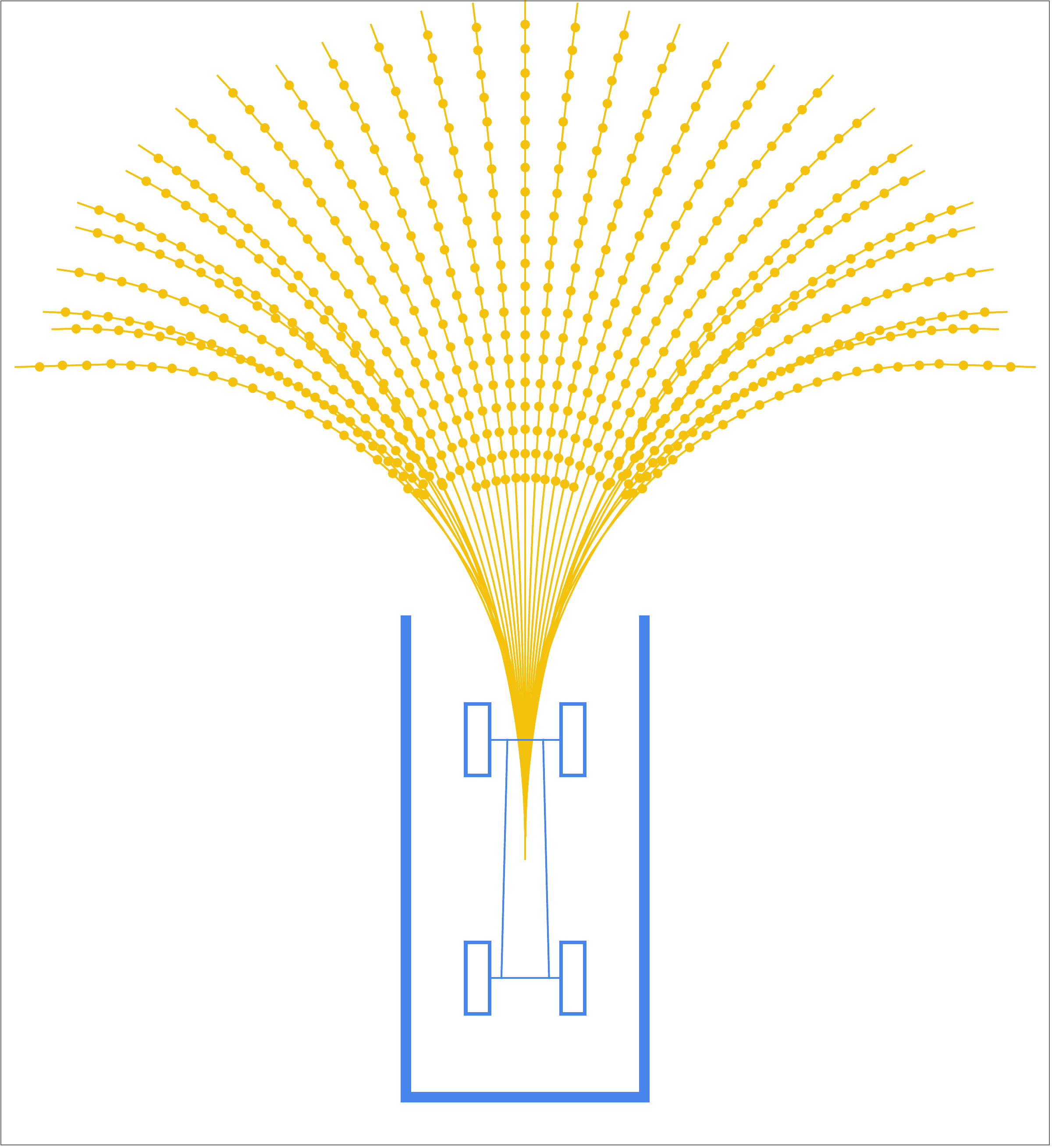}
	}
	\subfloat[Echelon]{
		\includegraphics[width=0.22\textwidth]{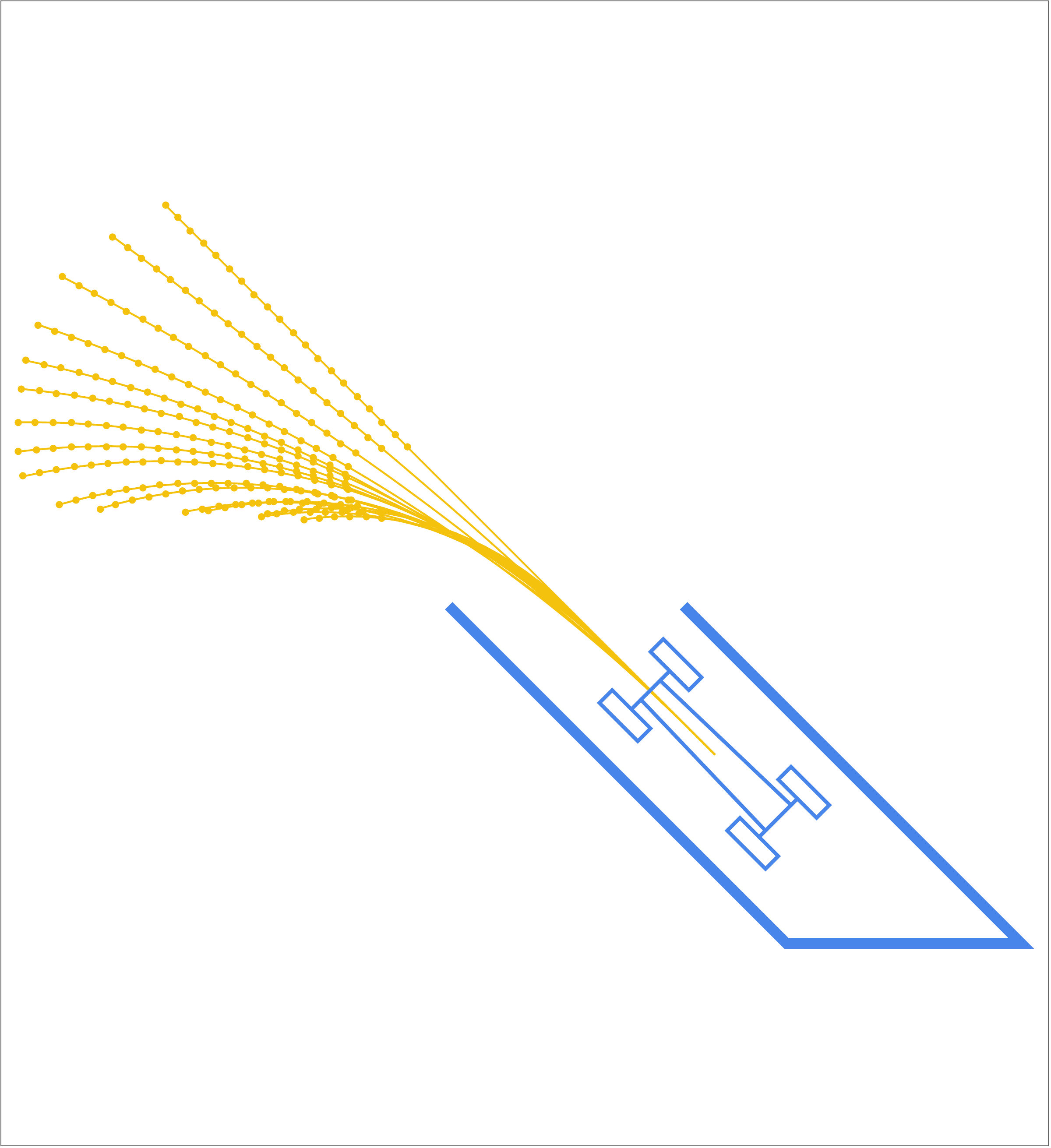}
	}\\
	\subfloat[Parallel]{
		\includegraphics[width=0.455\textwidth]{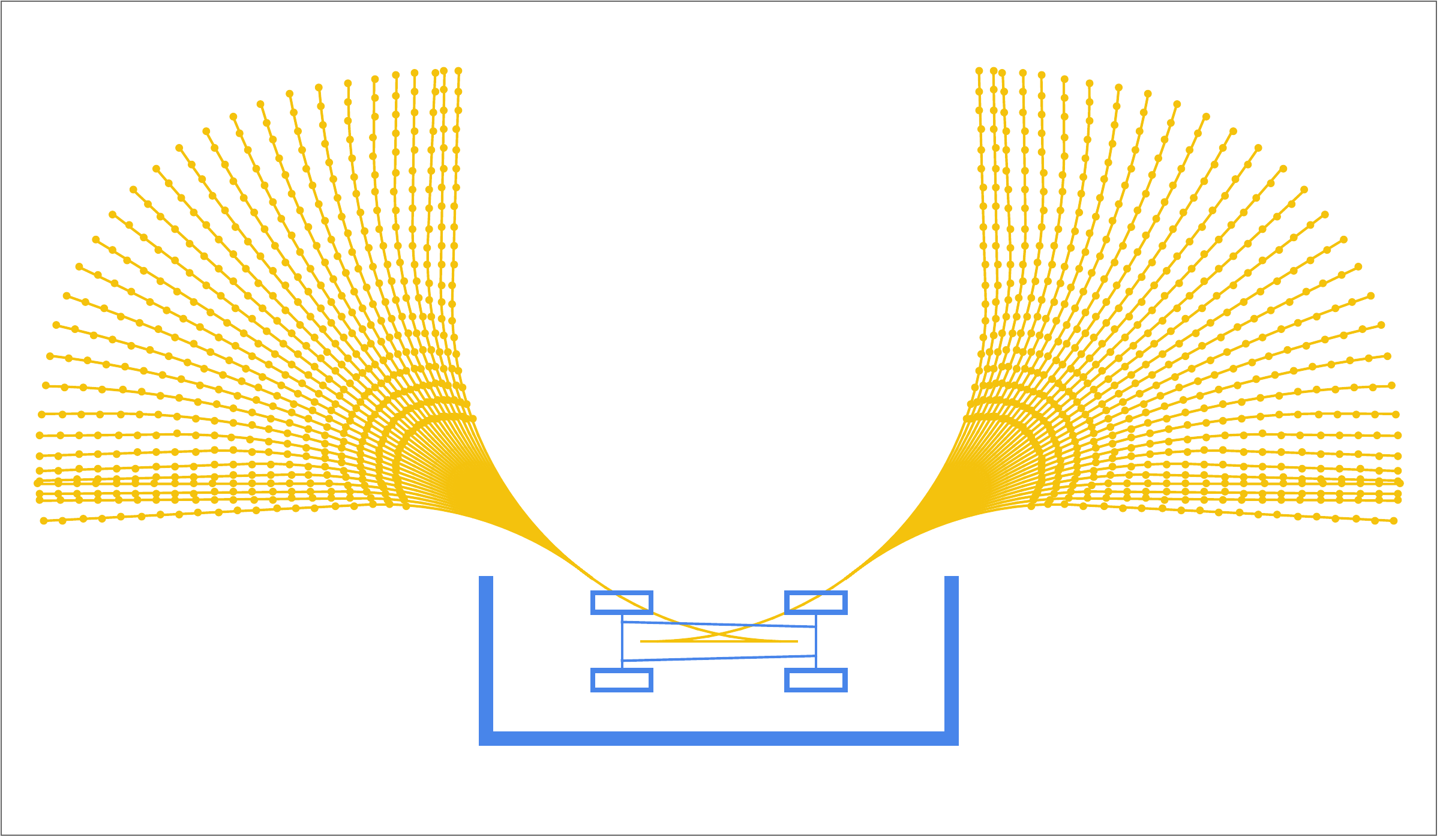}
	}
	\caption{Model-based target tree with dozens of paths and hundreds of nodes.}
	\label{fig:tree}
\end{figure}

In the basic RRT algorithm, when the distance between the car and the target is small but still larger than tolerance value $\theta$, it takes a great amount of effort to reduce the distance even slightly. The generated trajectory will be complex, and the car will move forward and backward multiple times to adjust to a perfect position. In order to solve this problem, we put forward a model-based target, which expands the target point to a model-based tree. As shown in Fig. \ref{fig:tree}, this model-based tree has hundreds of nodes. The RRT tree needs to get close to anyone of the target tree's nodes. Then we connect the paths from RRT tree and target tree to generate the final trajectory. This expansion enables the algorithm to finish its task more easily and makes the obtained trajectory smoother. The last part of trajectory will be perfect because the target tree is generated by the well-defined model. The car will move to target position without any adjustment.

To implement this algorithm, we need to generate a target tree and redefine the basic RRT's `random choice' function. It is shown in Algorithm \ref{alg3}.

To generate the target tree, we divide the parking scenario into three groups: perpendicular parking, parallel parking and echelon parking. Each model is made up of a few pre-defined parking routes. To generate the models, we imagine the car is already parked within the parking lot, and then we try to drive the car out of the parking lot with various pre-defined driving strategies. After this operation, we will get dozens of feasible driving-out paths, and each path has twenty nodes. Reverse these driving-out paths to parking-in paths, there will be a tree with dozens of paths and hundreds of nodes.

\begin{itemize}
	\item Perpendicular parking:
	Consider a car driving out of a perpendicular parking lot, it will go straight for sometime and then keep to a fixed turning angle. Since this angle changes from $-30^{\circ}$ to $30^{\circ}$ per $2^{\circ}$, we will get 31 different lines. Under the premise of a collision-free parking process, we will set the straight part as short as possible. Finally, as we pick 20 points per line as nodes of the model, we will get 620 nodes in the perpendicular model.
	
	\item Parallel parking:
	A car is driving out of a parallel parking lot, it will move backward for a while and then keep a max turning angle until its body is half out of the parking lot. Then it will choose a fixed turning angle from $-30^{\circ}$ to $30^{\circ}$ per $2^{\circ}$ at the rest of the path. In this part, we will also get 31 lines and 620 nodes, but considering that a car parking in a parallel parking lot would have two possible facing directions, the obtained data in our model is doubled up to 62 lines and 1240 nodes.
	
	\item Echelon parking:
	Similar to perpendicular parking, a car in an echelon parking lot will go straight for a while and then keep a fixed turning angle. The difference is that in this circumstance the whole movement is backward. The fixed turning angle changes from $0^{\circ}$ to $30^{\circ}$ per $2^{\circ}$. we will finally get 16 lines and 320 nodes.

\end{itemize}
\begin{algorithm}
	\caption{Random Choice}
	\label{alg3}
	\begin{algorithmic}
		\State $ r \  \gets \  random$($0 $,$  1 $)$ $
		\If{$r \  < \  0.5  $}
		\State \Return $ \  random\_point\_in\_free\_space$
		\Else
		\State \Return $ \  nodes\_in\_target\_tree $($ in \  proper \  order $)$ $
		\EndIf
	\end{algorithmic}
\end{algorithm}

\subsection{Smoothing under Kinematic Constraint}

Because of the random sampling of RRT tree nodes, there will always be some redundant nodes making the path complex. Therefore, we need an algorithm that can smooth it. There are already some smoothing algorithms for this problem, such as cubic polynomials \cite{cubic}, quintic polynomials \cite{quintic}, \cite{quintic2}, Bezier curves \cite{bezier}, \cite{bezier3}, \cite{bezier4}, B-splines \cite{bsp} and Clothoids \cite{clothiods}. Long Han \emph{et al.} also proposed a path smoothing algorithm \cite{2011RRT} for RRT-based parking planning. Both of these algorithms ignored the kinematic constraints of a car, which may cause the smoothed path to be unfeasible. By contrast, we have considered and worked on the vehicles kinematic constraints in our proposed smoothing algorithm, which ensures the generation of a feasible path. Our algorithm is based on the idea of Divide and Conquer. Firstly, for example, if the path has N nodes, we try to use `point pursuit' algorithm within N steps to find a new path from start point to target point. If success, the new path is shorter than the original path. If not successful, we will divide the path into two paths from the mid-node and execute the same process to each path until success or they will have just one node. After connecting every resulting path, we will get a smoother path. The `point pursuit' algorithm is under the car's kinematic constraints, so the generated path is feasible.

\begin{algorithm}
	\caption{Smooth}
	\label{alg4}
	\begin{algorithmic}
		\Function{smooth}{$path$}
		\If{$path \  just \  have \  one \  node$}
		\State \Return $path$
		\Else
		\State $new\_path.add(path.start)$
		\While{$new\_path\ is\ shorter\ than\ path$}
		\State $new\_path.add(\ point\_pursuit\ to\ path.end\ )$
		\If{$new\_path\ reached\ path.end$}
		\State \Return $new\_path$
		\EndIf
		\EndWhile
		\State $path\_1 $,$ \  path\_2 \gets divide$($path$)$ $
		\State $path\_1 \gets smooth$($path\_1$)$ $
		\State $path\_2 \gets smooth$($path\_2$)$ $
		\State \Return$connect$($path\_1 $,$  path\_2$)$ $
		\EndIf
		\EndFunction
	\end{algorithmic}
\end{algorithm}

\subsection{Parking Planning}

As for parking planning, we modified the basic RRT algorithm with the model-based tree and smooth algorithm previously mentioned. Firstly, our algorithm will assign the parking task to one of three parking scenarios, and then corresponding functions are called to grow the target tree. A list of nodes in the target tree will be sent to the Algorithm \ref{alg3}. Finally, we will run the RRT algorithm to generate a parking path. In every step of RRT, the Algorithm \ref{alg3} will choose a target point for it. RRT will check if the bias is under the tolerance value after every 1000 steps, which means that the RRT will run at least 1000 steps. If the bias is under the tolerance value, the RRT algorithm is terminated and we will start the smoothing algorithm to smooth the path before outputting it.
\begin{algorithm}[hpb]
	\caption{Parking Planning}
	\label{alg5}
	\begin{algorithmic}
		
		\State $target\_tree \gets model\_based\_tree$($ $)$ $
		\State $node\_list\gets target\_tree.nodes$
		\State $Tree\  T\gets start\_Point$
		\While{$distance$($T,target\_point$)$>tolerance\_value $}
		\For{$1000\ times$}
		\State $P\gets random\_choice$($node\_list$)$ $
		\State $N\gets nearest\_node\_in\_tree$
		\State $E,\  A\gets point\_pursuit$($N,P$)$ $
		\State $T.add$($E, A$)$ $
		\EndFor
		\EndWhile
		\State $RRT\_path\gets T.backtrack$($ $)$ $
		\State $RRT\_path\gets smooth$($RRT\_path$)$ $
		\State $model\_path\gets target\_tree.backtrack$($ $)$ $
		\State $path\gets RRT\_path+model\_path$
		\State \Return $path$
		
	\end{algorithmic}
\end{algorithm}

\begin{figure}[tb]
	\centering
	\subfloat[Perpendicular]{
		\includegraphics[width=0.22\textwidth]{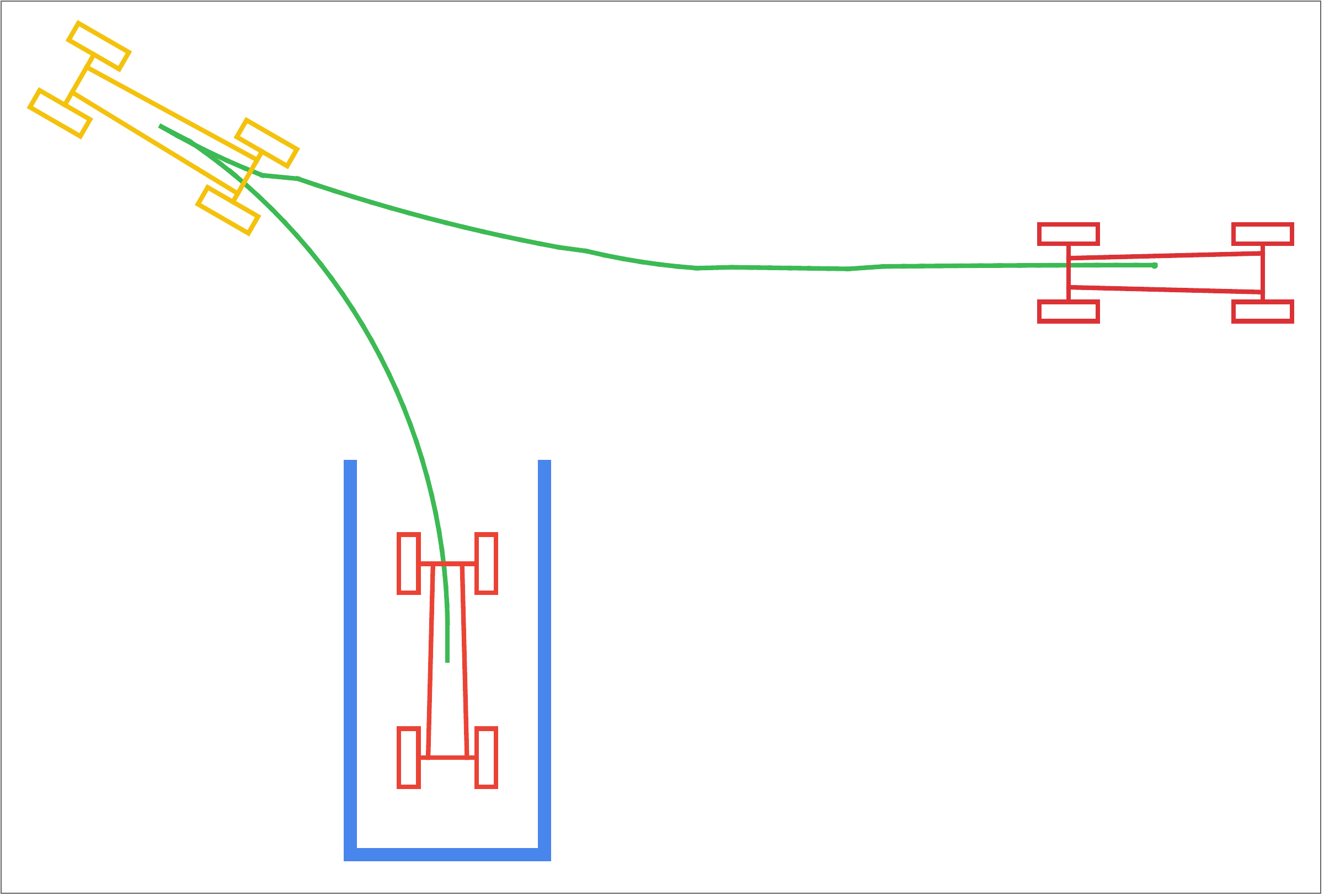}
	}
	\subfloat[Echelon]{
		\includegraphics[width=0.22\textwidth]{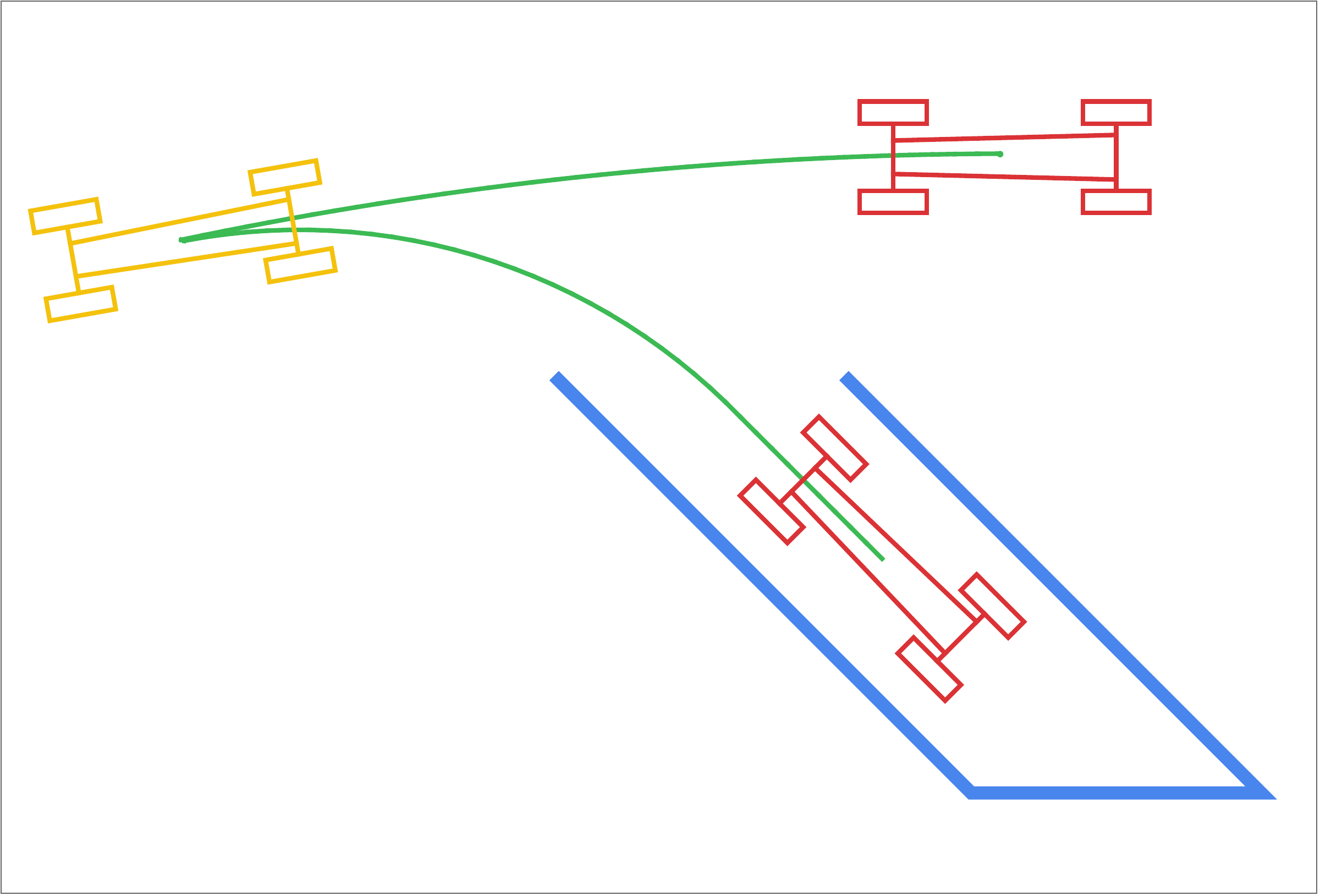}
	}\\
	\subfloat[Parallel]{
		\includegraphics[width=0.455\textwidth]{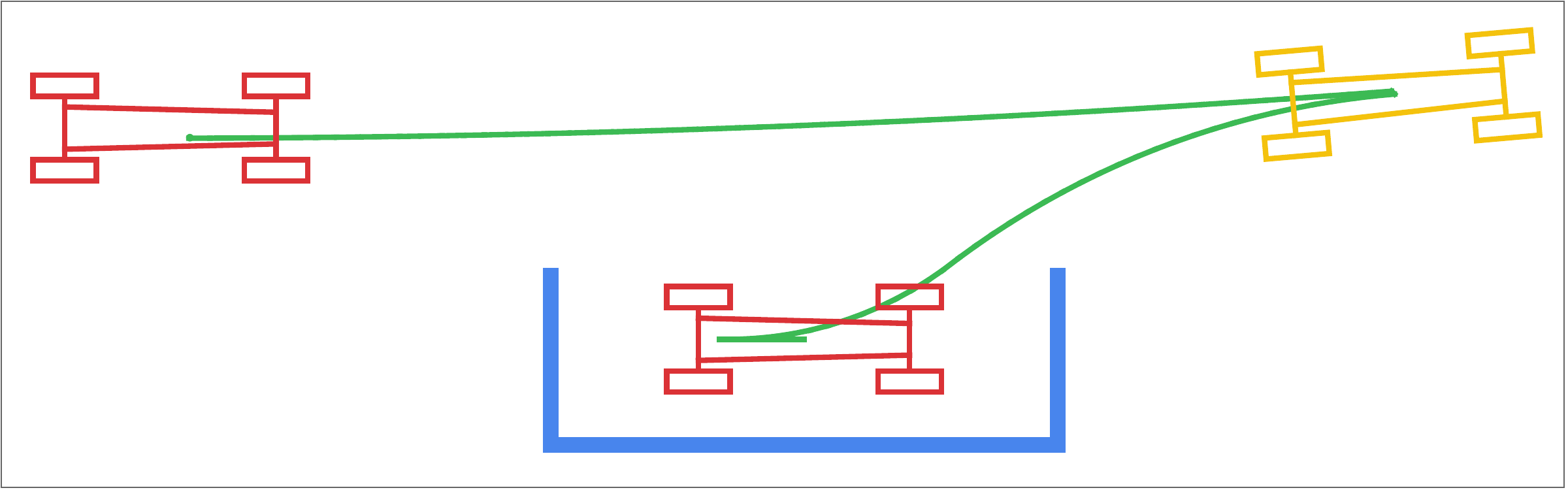}
	}
	\caption{A parking path generated by our algorithm in three different parking scenarios.}
	\label{fig:ourpath}
\end{figure}

\begin{figure}[tb]
	\centering
	\subfloat[Basic RRT]{
		\includegraphics[width=0.15\textwidth]{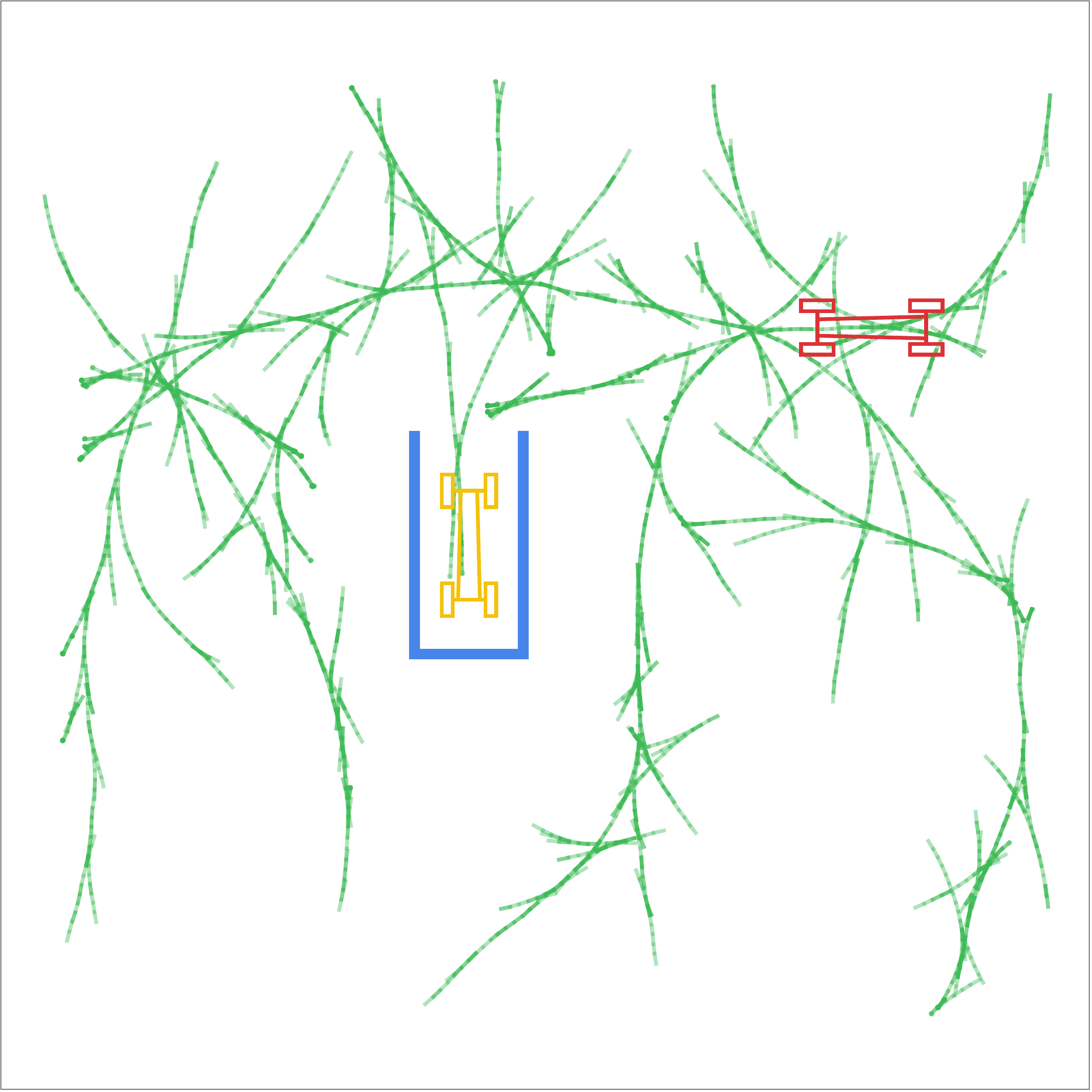}
	}
	\subfloat[Basic RRT]{
		\includegraphics[width=0.15\textwidth]{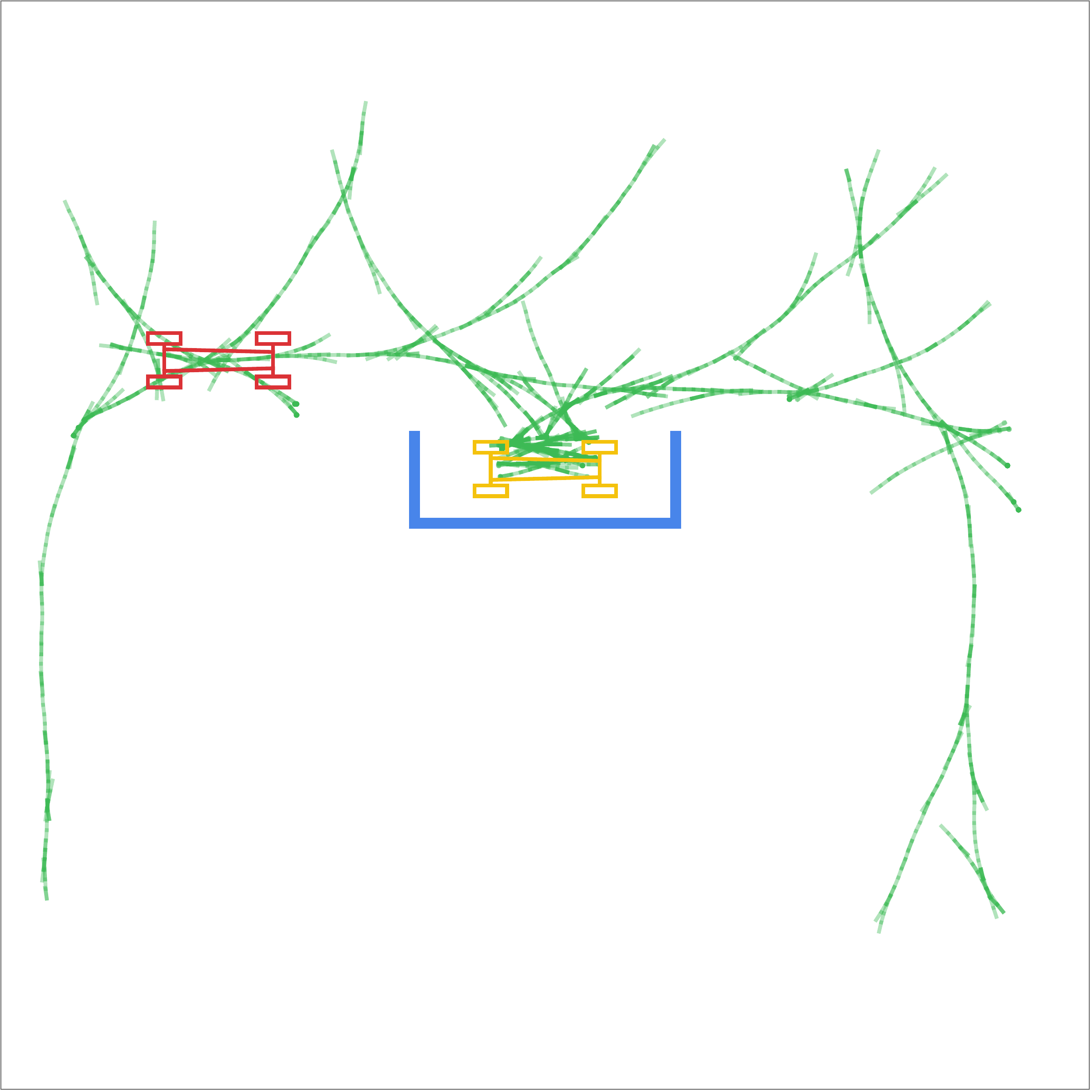}
	}
	\subfloat[Basic RRT]{
		\includegraphics[width=0.15\textwidth]{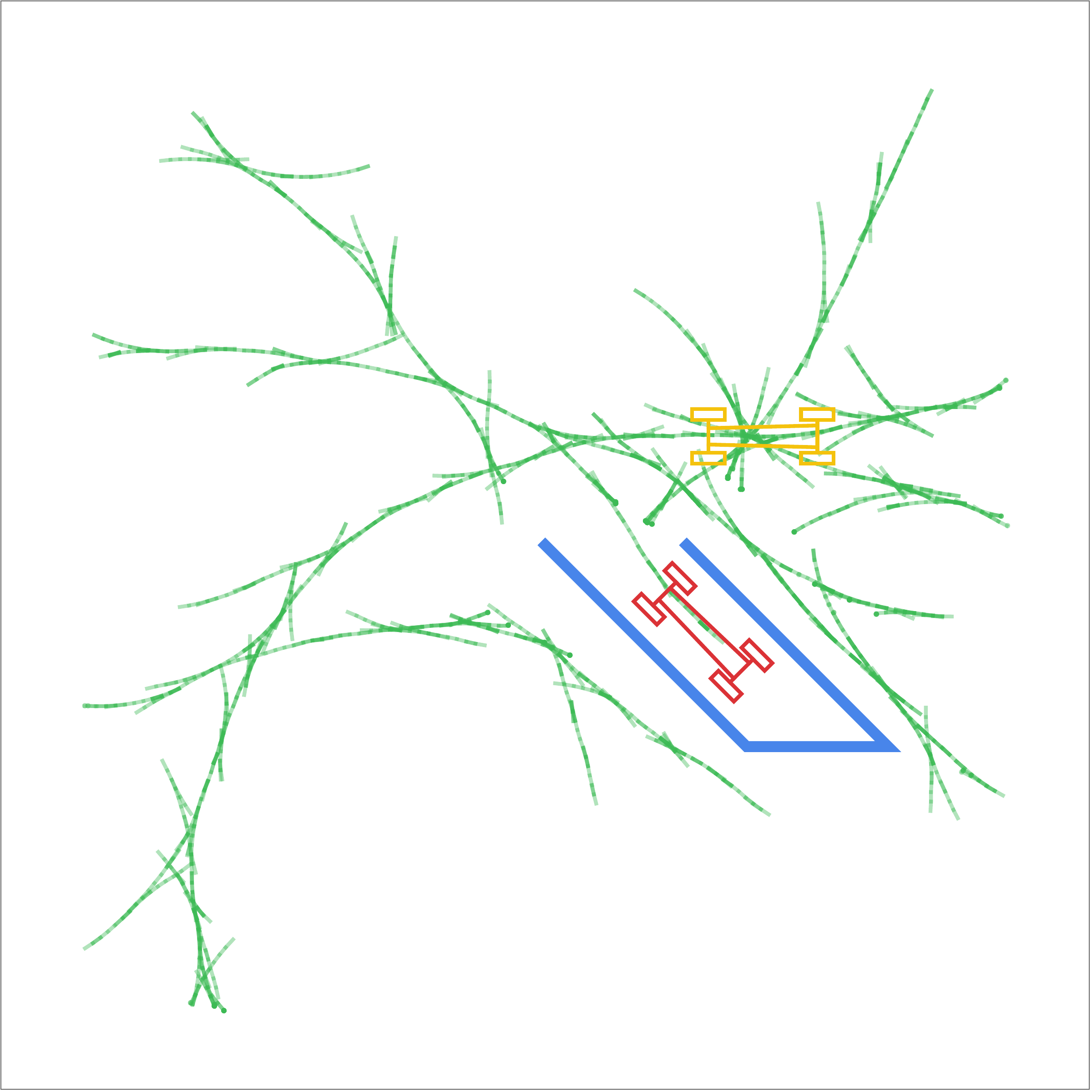}
	}\\
	\subfloat[Proposed method]{
		\includegraphics[width=0.15\textwidth]{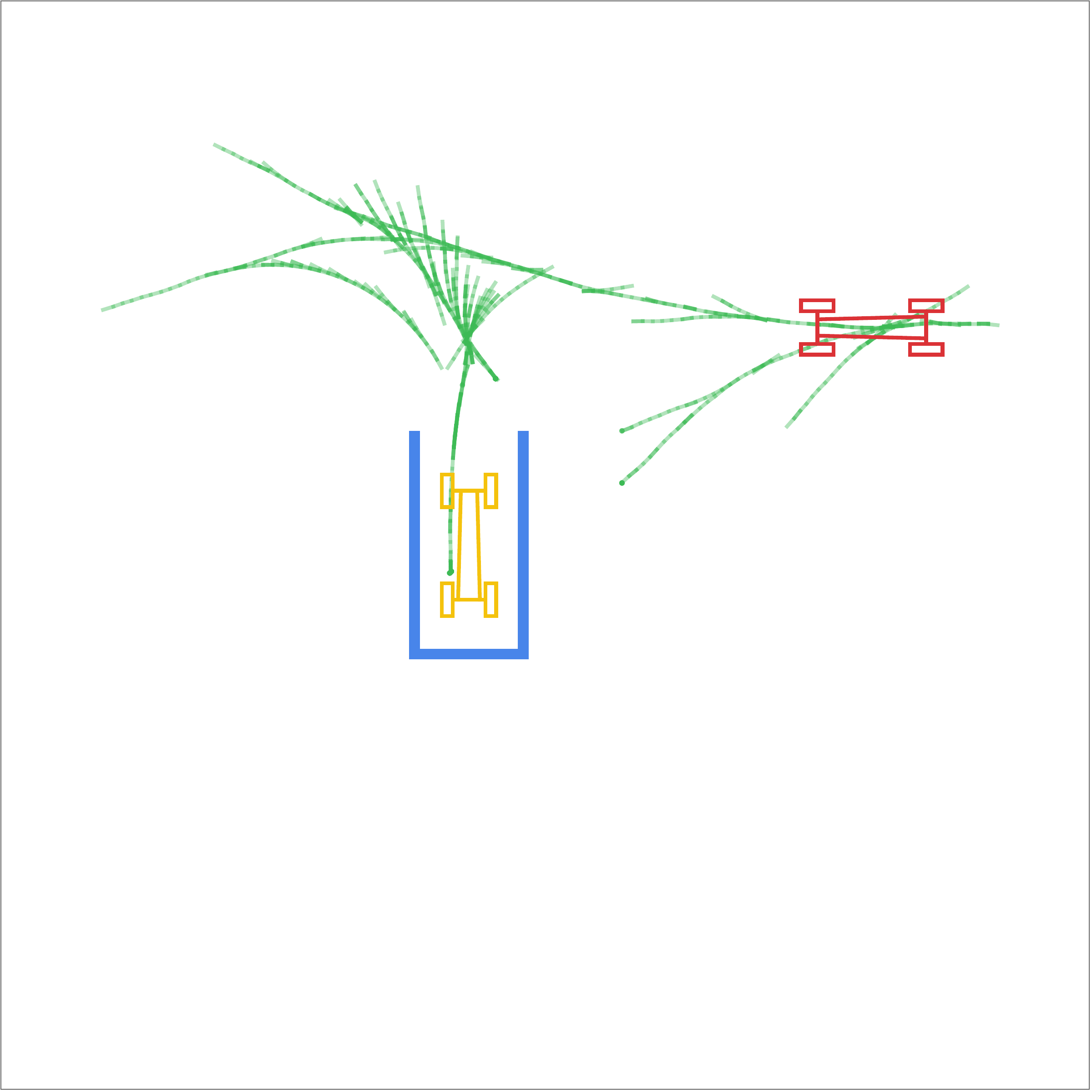}
	}
	\subfloat[Proposed method]{
		\includegraphics[width=0.15\textwidth]{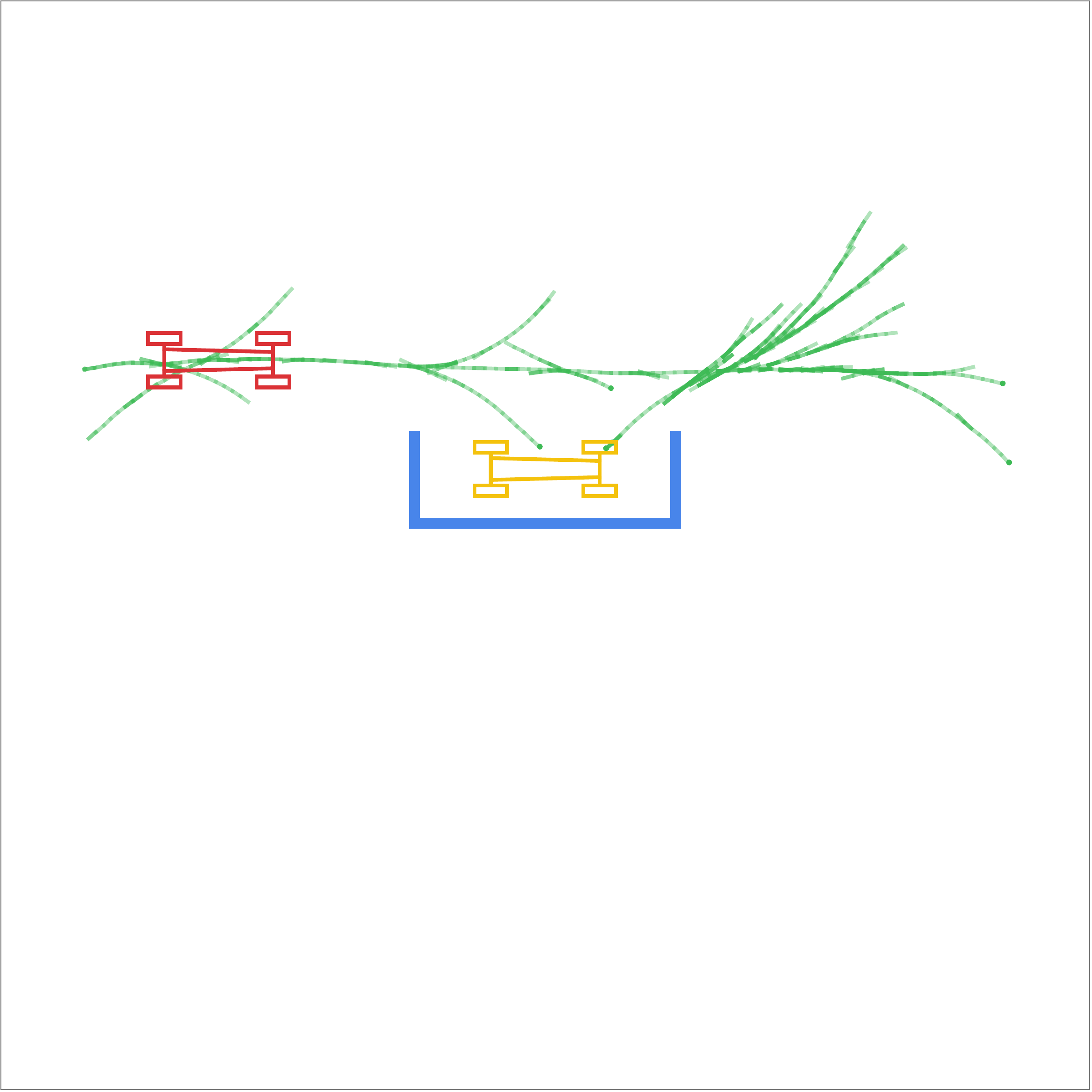}
	}
	\subfloat[Proposed method]{
		\includegraphics[width=0.15\textwidth]{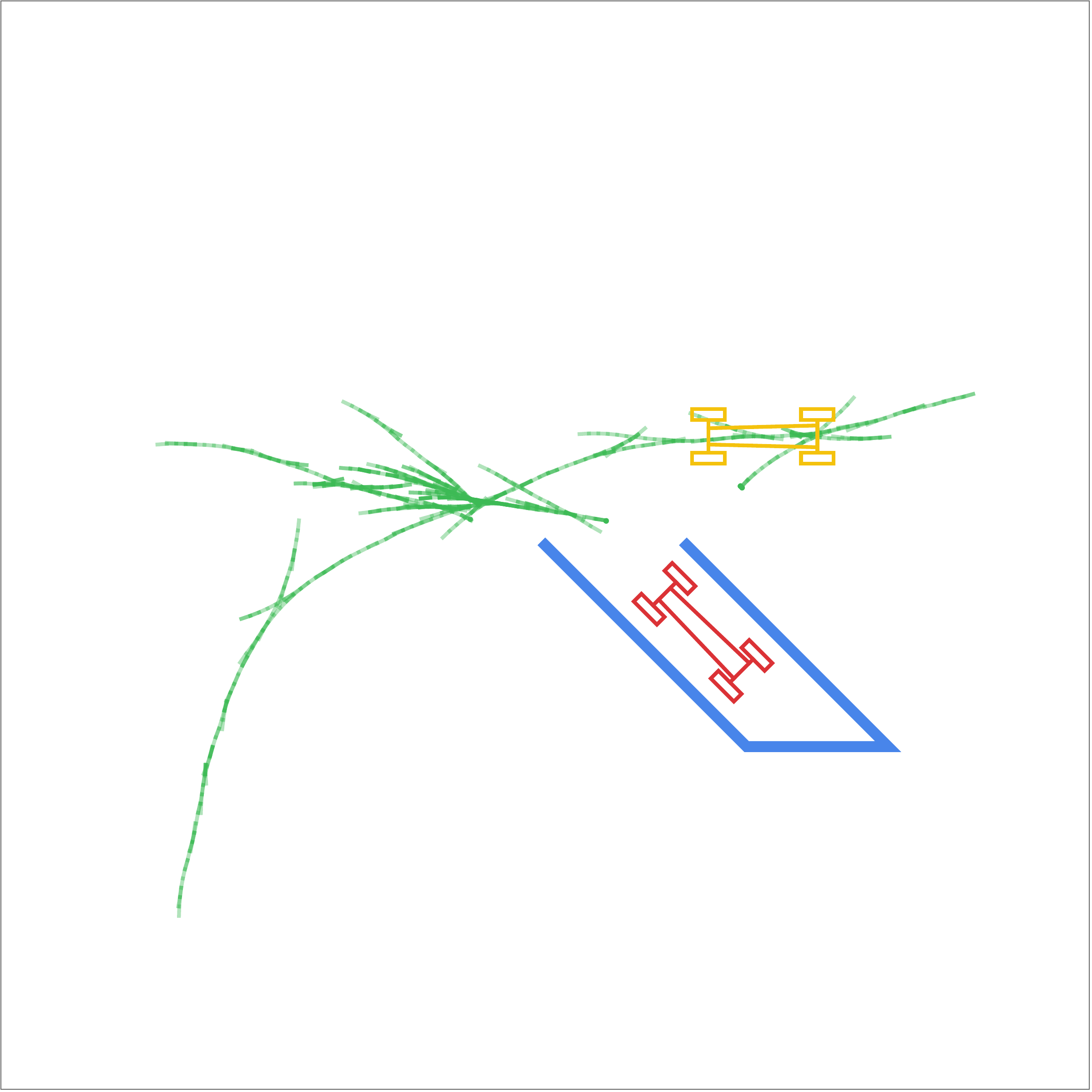}
	}
	\caption{Searched space in three different parking scenarios with two algorithms. (a), (b), (c) is the searched space of basic RRT; (d), (e), (f) is the searched space of our algorithm. This searched space doesn't include the branches in target tree.}
	\label{fig:search_space}
\end{figure}

\begin{figure}[tb]
	\centering
	\subfloat[Basic RRT]{
		\includegraphics[width=0.15\textwidth]{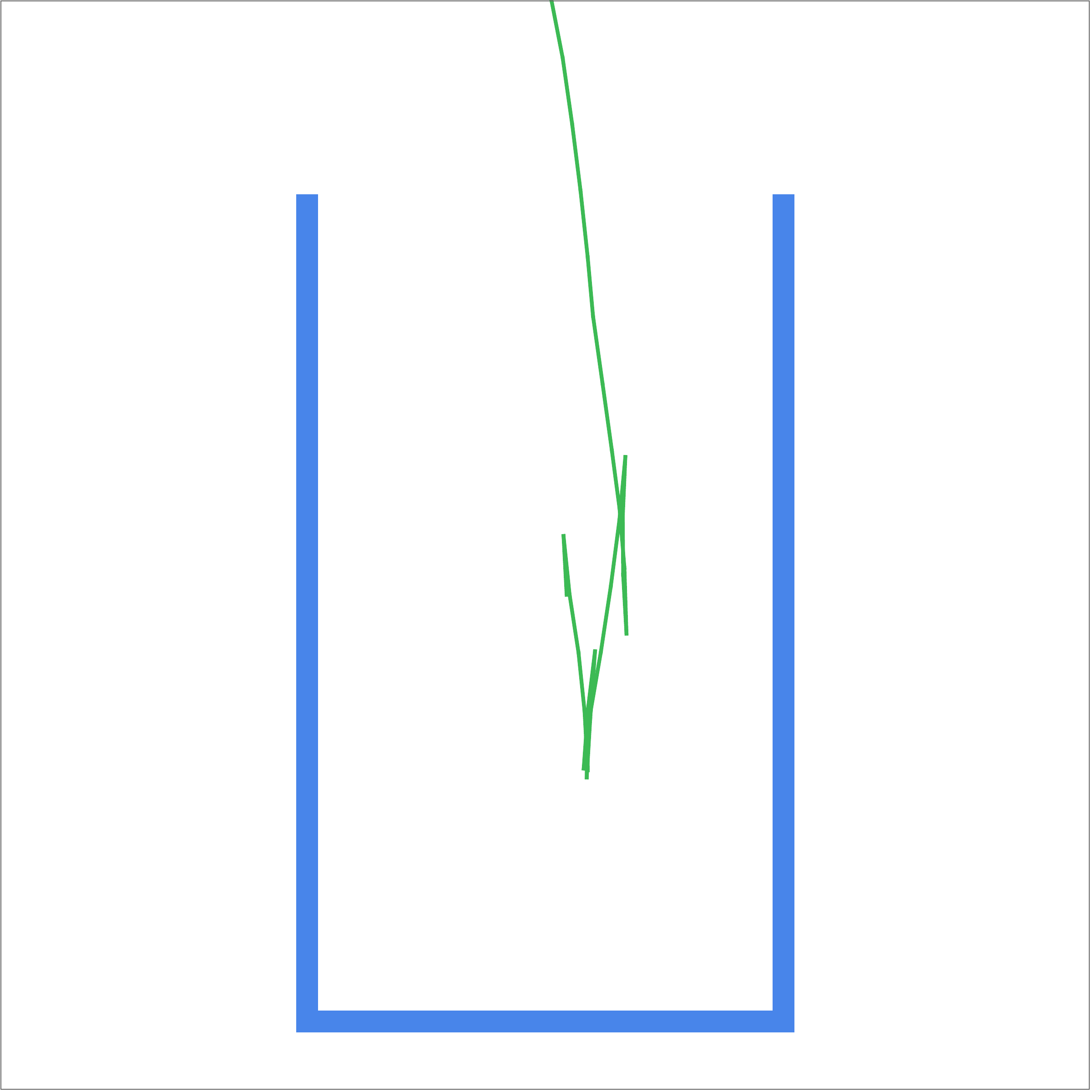}
	}
	\subfloat[Basic RRT]{
		\includegraphics[width=0.15\textwidth]{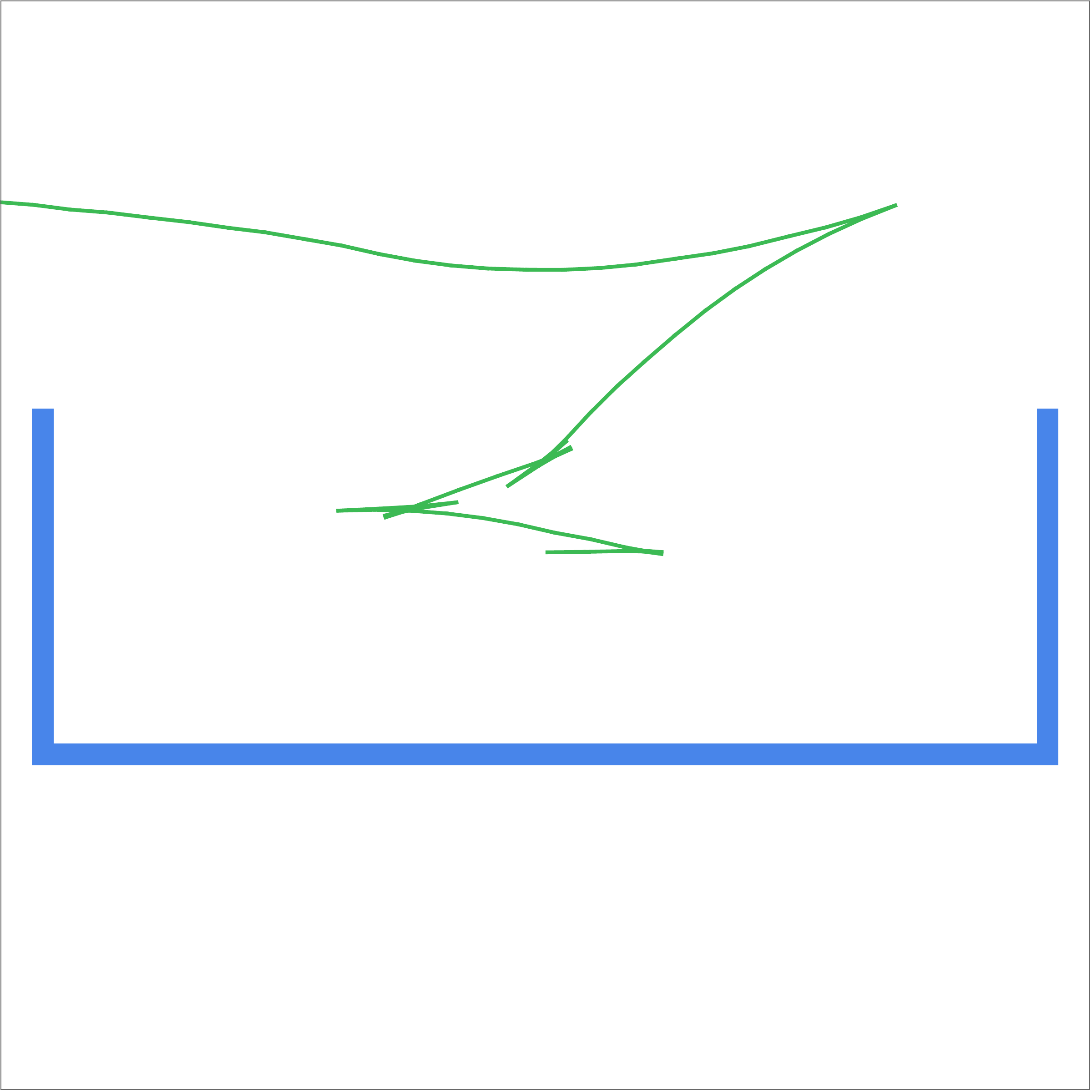}
	}
	\subfloat[Basic RRT]{
		\includegraphics[width=0.15\textwidth]{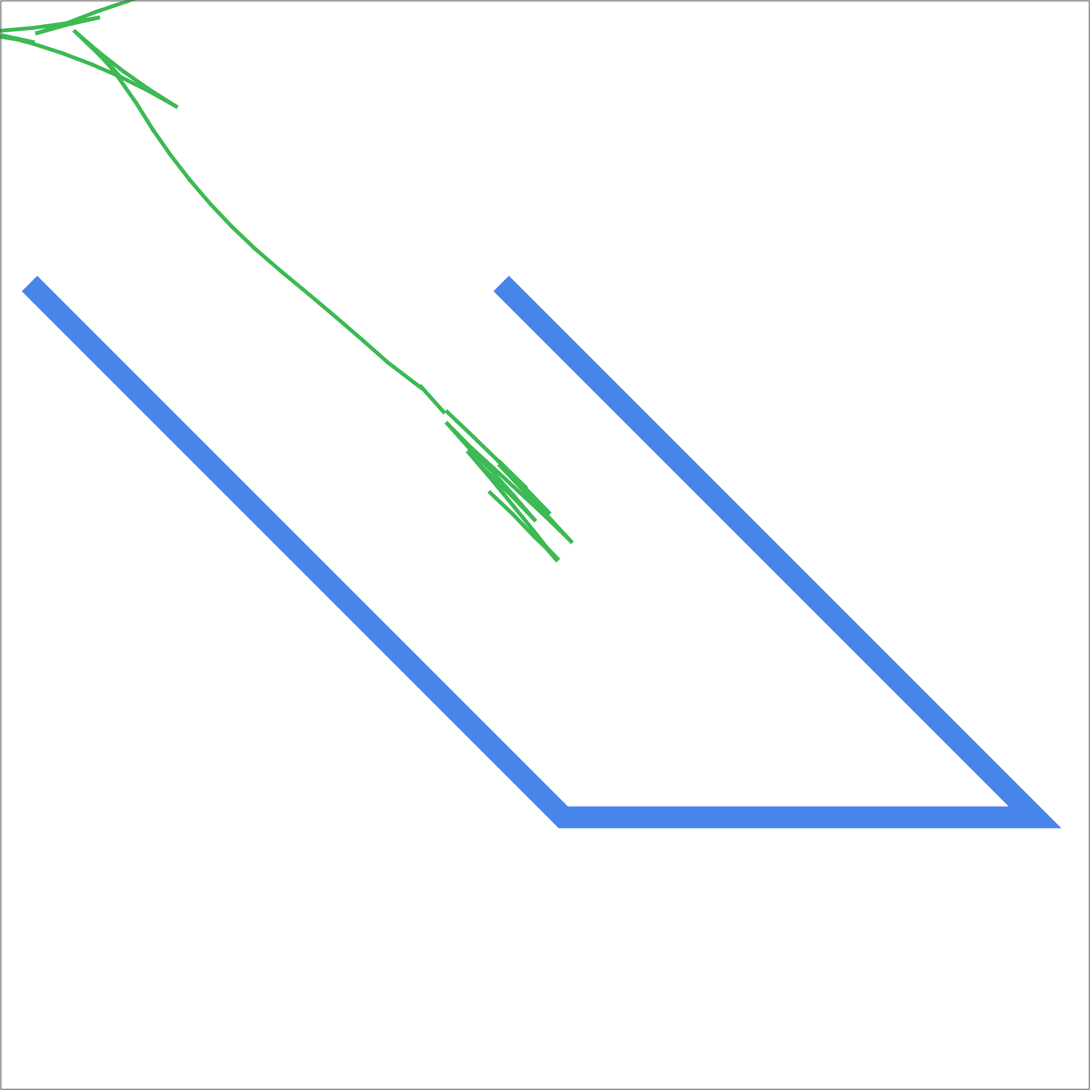}
	}\\
	\subfloat[Proposed method]{
		\includegraphics[width=0.15\textwidth]{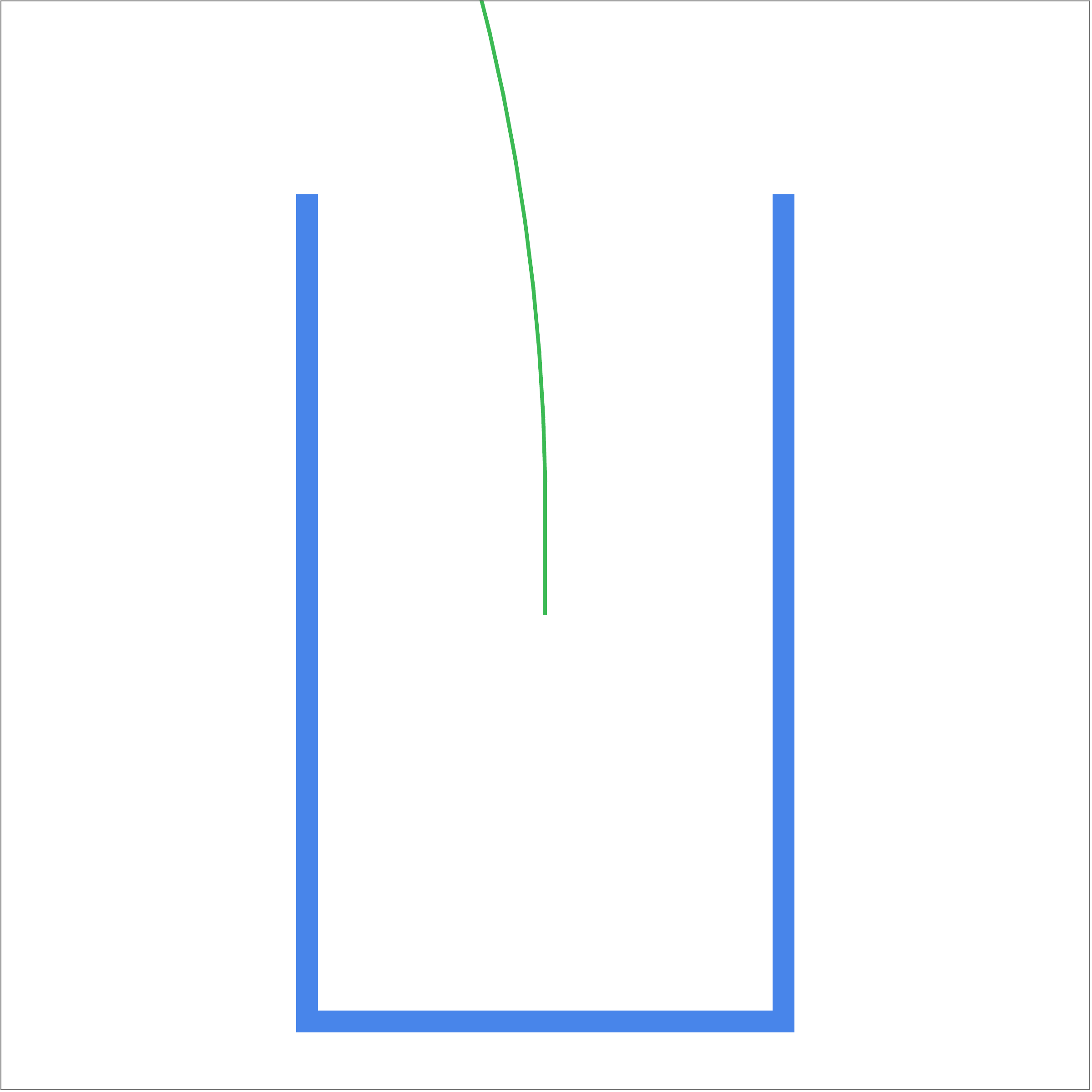}
	}
	\subfloat[Proposed method]{
		\includegraphics[width=0.15\textwidth]{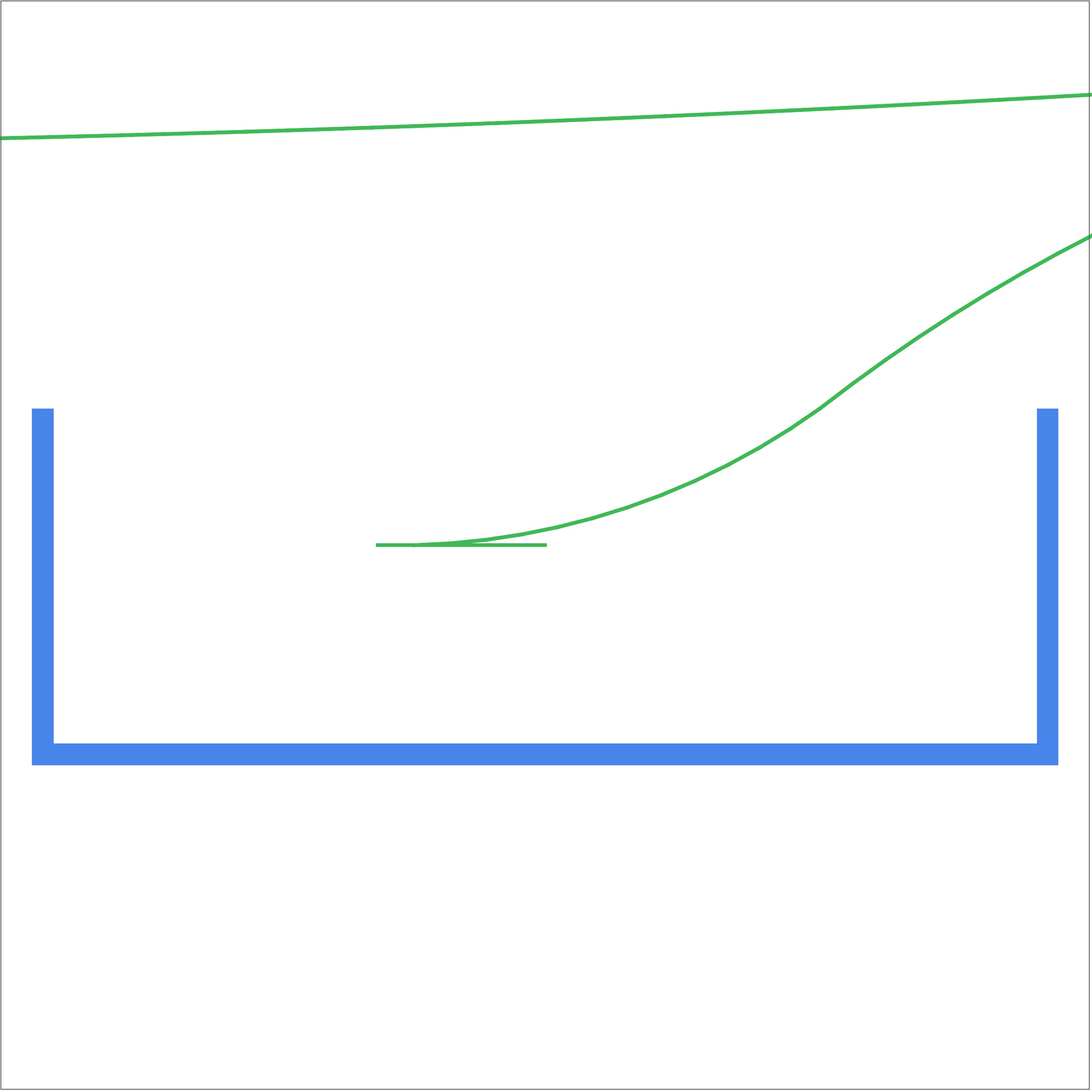}
	}
	\subfloat[Proposed method]{
		\includegraphics[width=0.15\textwidth]{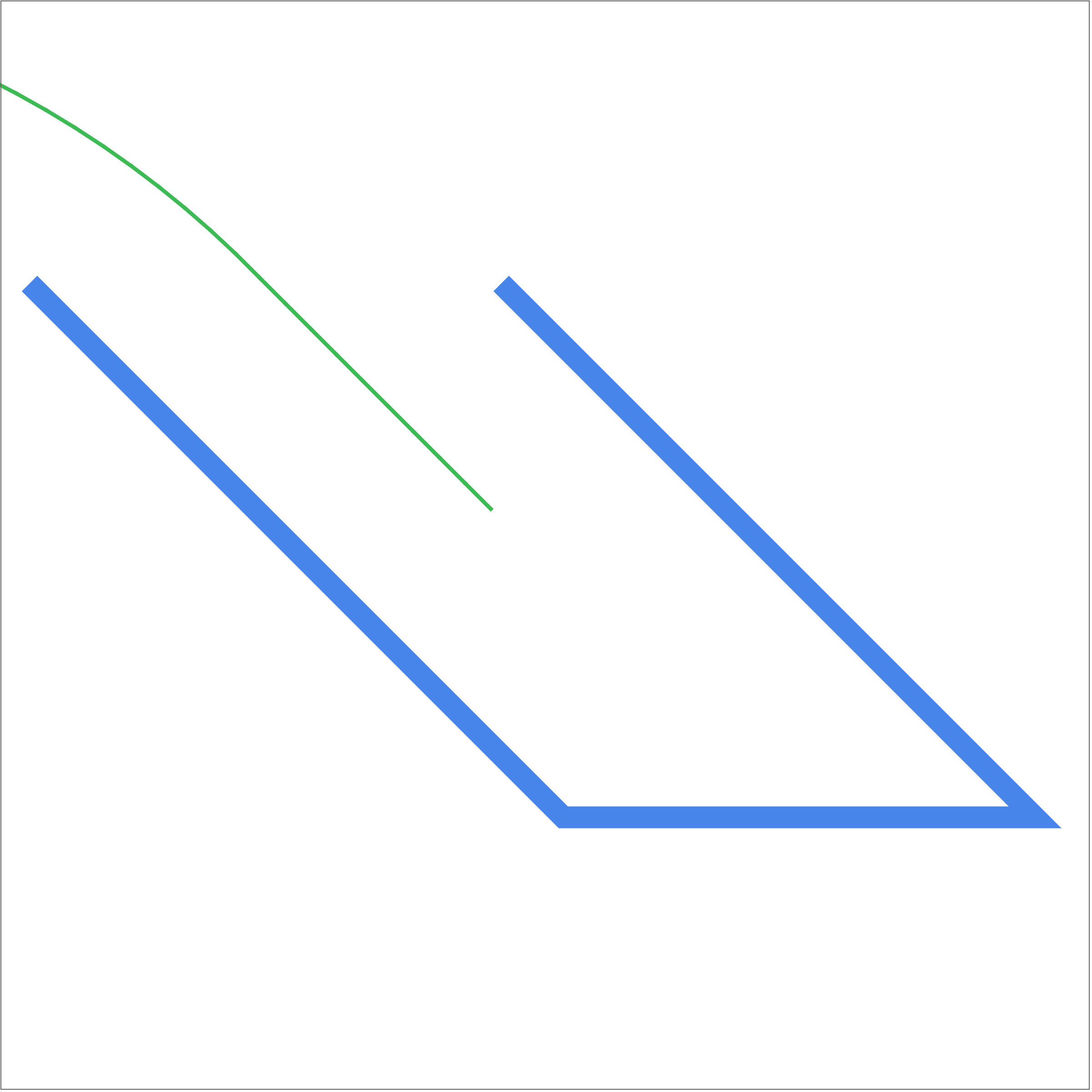}
	}
	\caption{The final part of the path. (a), (b), (c) is generated by basic RRT, which is very complex. The car will move forward and backward multiple times to adjust into a perfect position. (d), (e), (f) is generated by our algorithm. It is very smooth, the car can get to the target position in just one movement.}
	\label{fig:last_path}
\end{figure}

\begin{table*}
	\renewcommand{\arraystretch}{1.5}
	\caption{STEPS AND TIME COST}
	\label{table1}
	\centering
	\begin{tabular}{|c||c|c|c|c|c|c|}
		\hline
		Scenarios&\multicolumn{2}{c|}{Perpendicular Parking}&\multicolumn{2}{|c|}{Parallel Parking}&\multicolumn{2}{|c|}{Echelon Parking}\\
		\hline
		Target Type&Target Point&Model-based Target&Target Point&Model-based Target&Target Point&Model-based Target\\
		\hline
		Max Steps&9000&1000&5000&1000&13000&1000\\
		\hline
		Min Steps&4000&1000&2000&1000&6000&1000\\
		\hline
		Average Steps&6250&1000&3000&1000&8750&1000\\
		\hline
		Max Time&4.85s&0.20s&2.28s&0.23s&10.30s&0.20s\\
		\hline
		Min Time&1.46s&0.12s &0.13s&0.17s &2.98s&0.17s \\
		\hline
		Average Time&2.54s&0.17s&0.92s&0.20s&5.28s&0.18s\\
		\hline
	\end{tabular}
\end{table*}

\section{Experiment}
Our algorithm is programmed in C++ and executed on a personal computer with Ubuntu operating system. Our algorithm can generate the parking trajectory. We test the algorithm in three typical scenarios: perpendicular, parallel and echelon. The result is shown in Fig. \ref{fig:ourpath}. We will compare the result with basic RRT.

\subsection{Model-based Target}
In our experiment, there are two main advantages in the model-based target.
Firstly, the target tree extends the target point out of the parking lot obstacles and expands the target point from one point to hundreds of nodes, which makes the RRT tree easier and quicker in reaching its target. Then the RRT tree needs to search a smaller space with fewer steps to get close enough to the target tree. The searched space is shown in Fig. \ref{fig:search_space}. Generally, the more branches in the target tree, the smaller space needed to search, because of the reduction of search space and steps, our algorithm will have a speed boost. The steps and time cost in our algorithm compared to basic RRT is presented in Table \ref{table1}. After hundreds of tests, we confirmed that the basic RRT takes thousands of steps, and in some worse conditions, it takes even over ten thousands of steps. In comparison, our algorithm always takes 1000 steps and never fail. As previously mentioned, our algorithm will not stop before 1000 steps. Usually, the basic RRT costs 0.13s $\sim$ 10s for calculation, ours always takes about 0.2s. 

Secondly, the final part of our parking path is more smooth and the safe margin is bigger. The reason is as follows: every RRT algorithm has a connection zone in the path. The RRT tree will connect to the target point in the connection zone. In basic RRT, the connection zone is around the target point, while in our algorithm it is the place where the RRT tree connects to the model-based target tree. The final part of parking path in basic RRT includes the connection zone, resulting in it being extremely complex. The vehicle will move forward and backward multiple times inside parking lot to adjust to a suitable position. Since it is usually narrower inside the parking lot, the adjustment may cause a collision. The final part of our algorithm is generated from the well defined model, very smooth and human-like. The vehicle will move into a perfect position in just one step.

The last part of the path is shown in Fig. \ref{fig:last_path}. The connection zone of our algorithm is outside the parking lot, where it is much wider and safer. The target tree has hundreds of nodes, which makes it much easier to connect. The connection zone will also be very smooth. The connection zone of our algorithm is shown in Fig. \ref{fig:ourpath}.

\subsection{Smoothing with Vehicle Kinematic Constraints}
\begin{figure}[t]
	\centering
	\subfloat[Basic RRT]{
		\includegraphics[width=4cm]{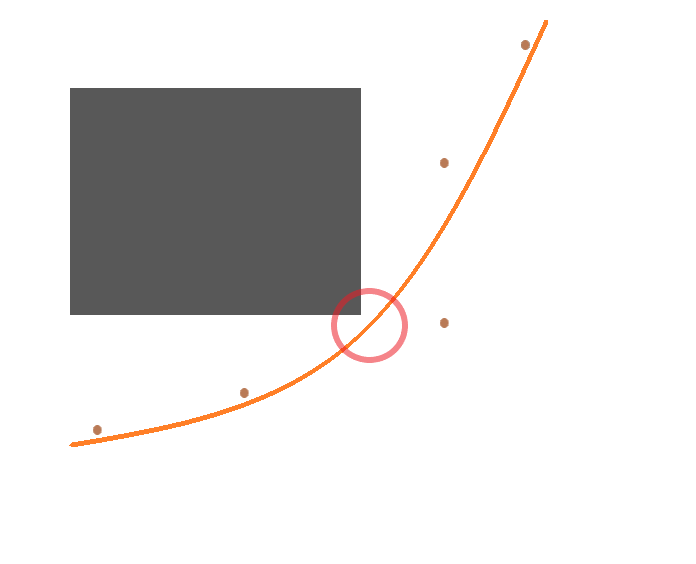}
	}
	\subfloat[Proposed method]{
		\includegraphics[width=4cm]{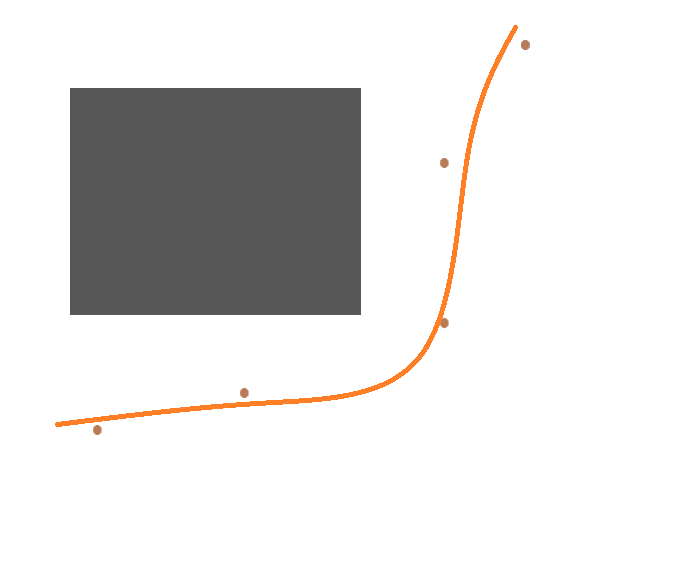}
	}\\
	\subfloat[]{
		\includegraphics[width=0.4\textwidth]{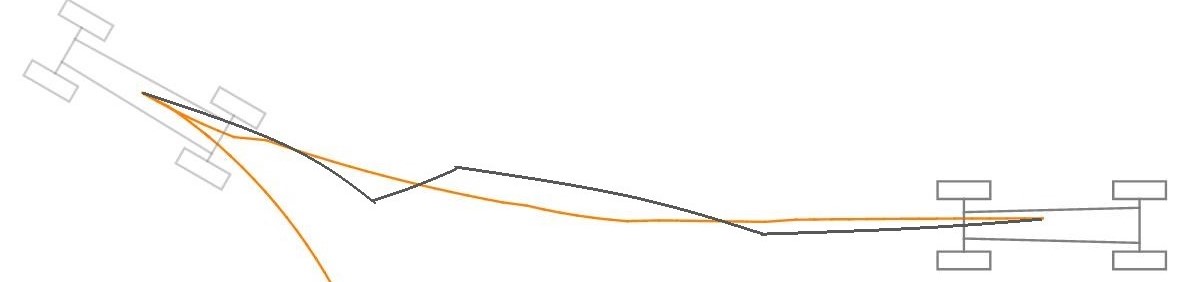}
	}\\
	\subfloat[]{
		\includegraphics[width=0.4\textwidth]{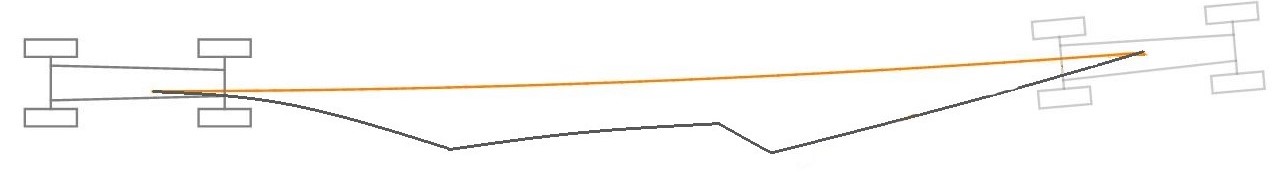}
	}\\
	\caption{Demo of smoothing algorithm. (a) shows the path smoothed by traditional smoothing algorithm. Violently distorting the path to smoothly curve results in collision in the red zone. (b) shows the path smoothed by our algorithm. In (c), (d), the black line is the path before smooth, and the orange line is the path smoothed by our smoothing algorithm.}
	\label{fig:comp_smo}
\end{figure}

\begin{figure}[t]
	\centering
	\subfloat[Before smooth]{
		\includegraphics[width=4cm]{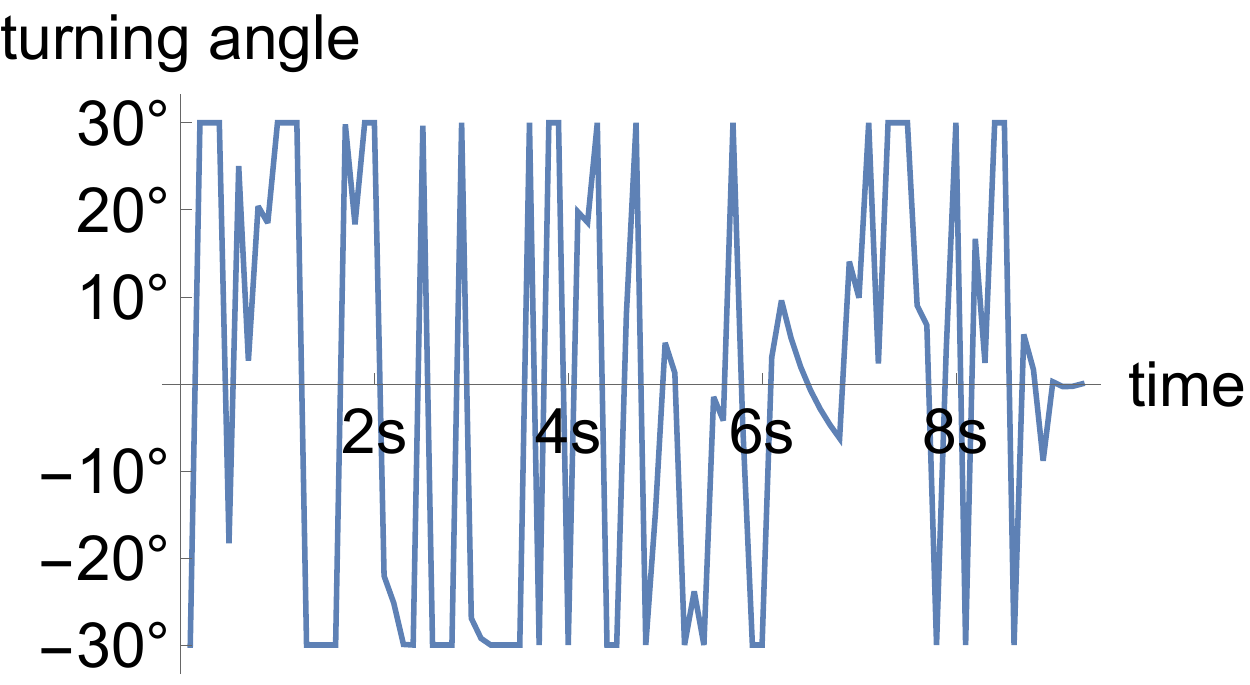}
	}
	\subfloat[Smoothed]{
		\includegraphics[width=4cm]{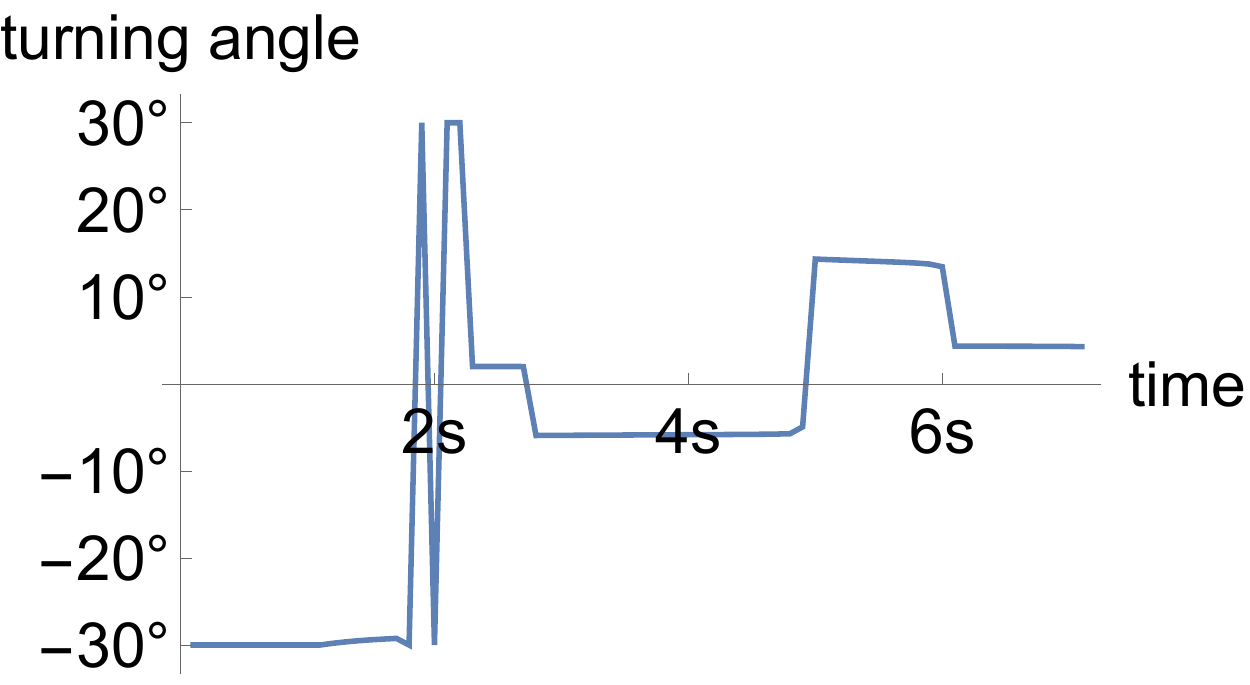}
	}\\
	\subfloat[Before smooth]{
		\includegraphics[width=4cm]{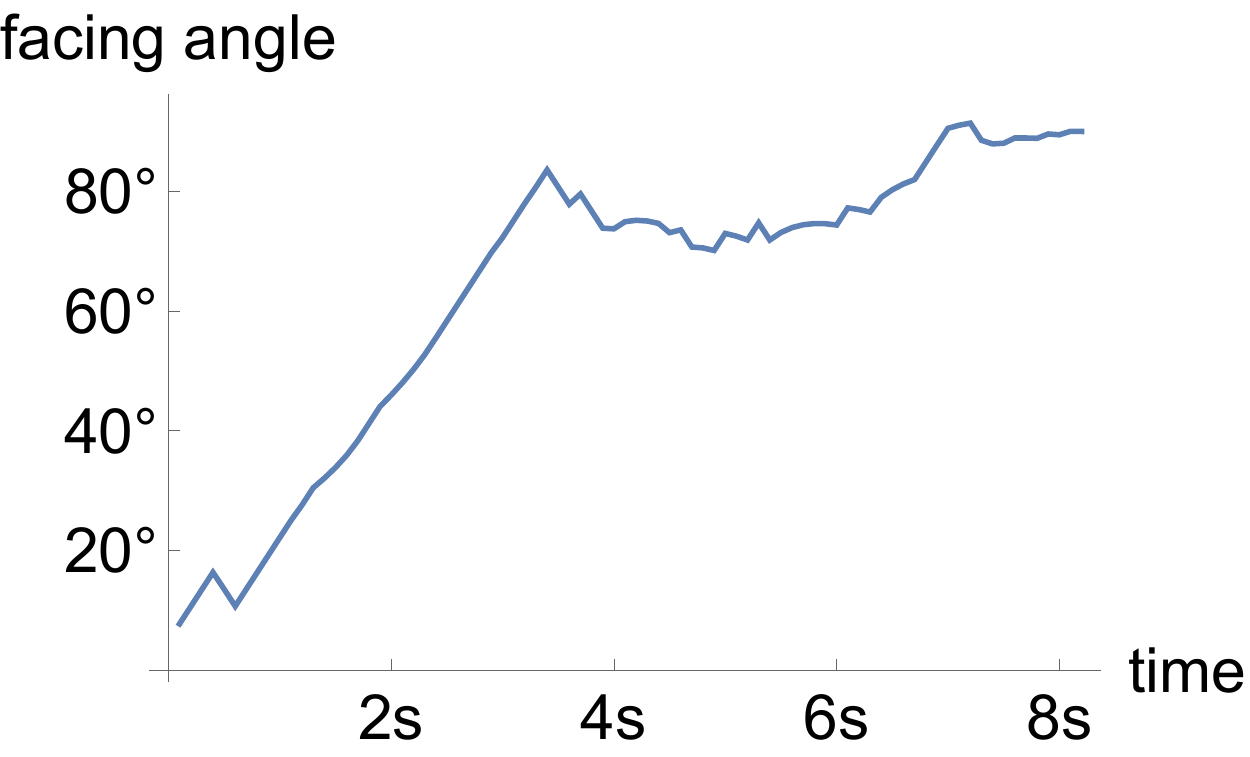}
	}
	\subfloat[Smoothed]{
		\includegraphics[width=4cm]{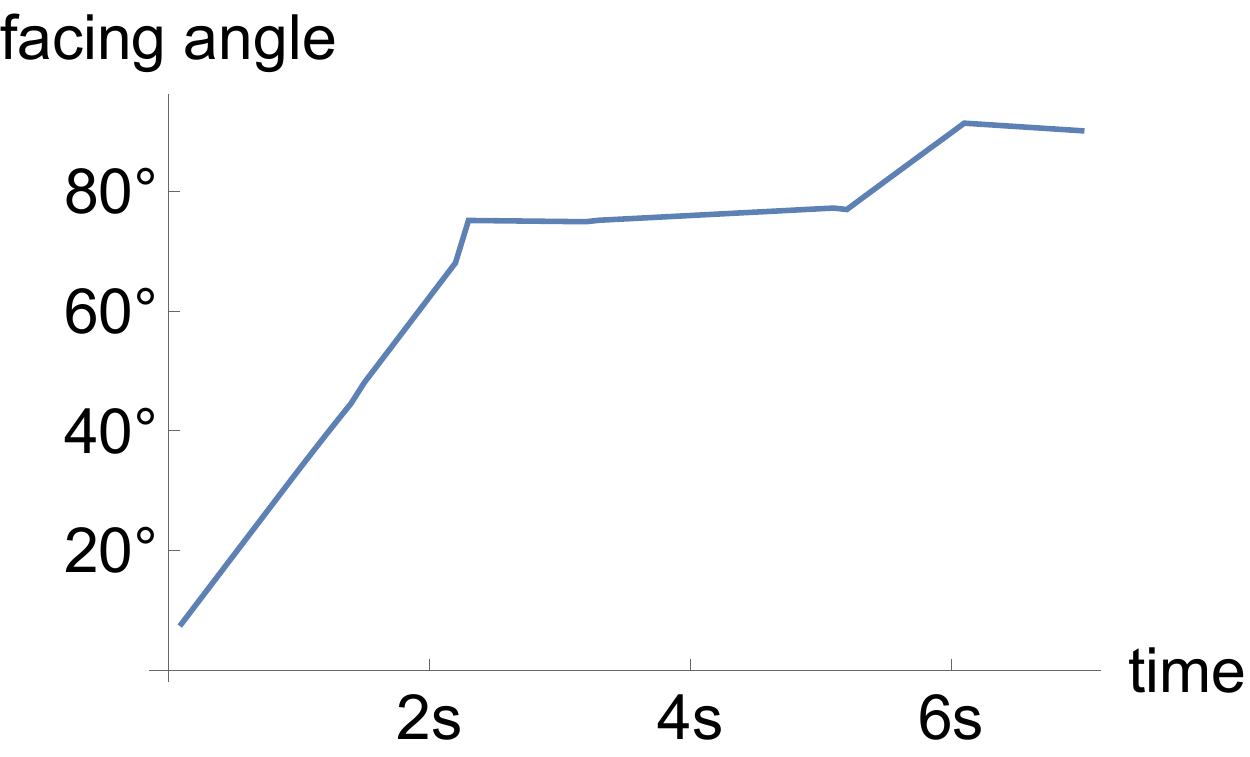}
	}\\
	\subfloat[Before smooth]{
		\includegraphics[width=4cm]{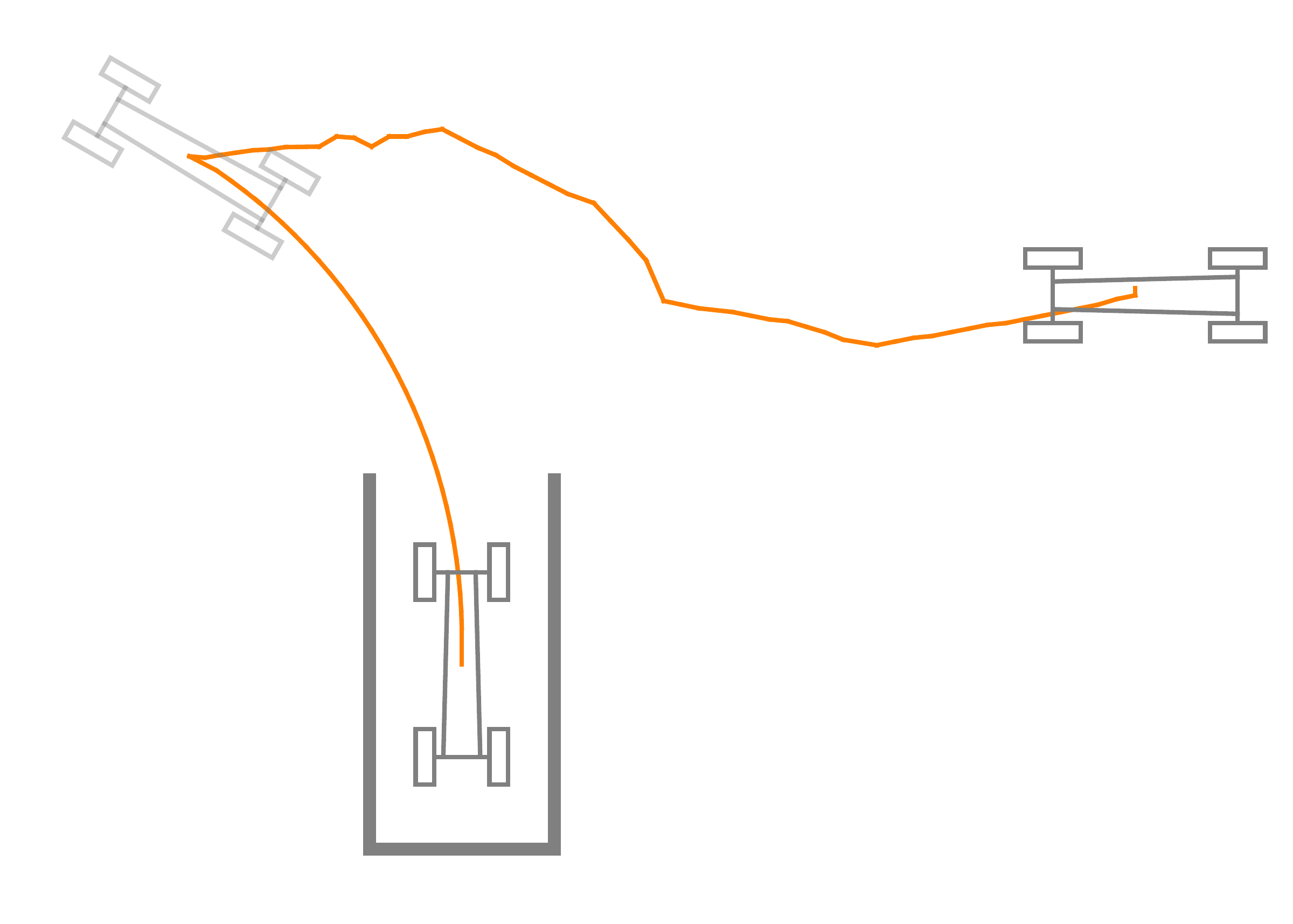}
	}
	\subfloat[Smoothed]{
		\includegraphics[width=4cm]{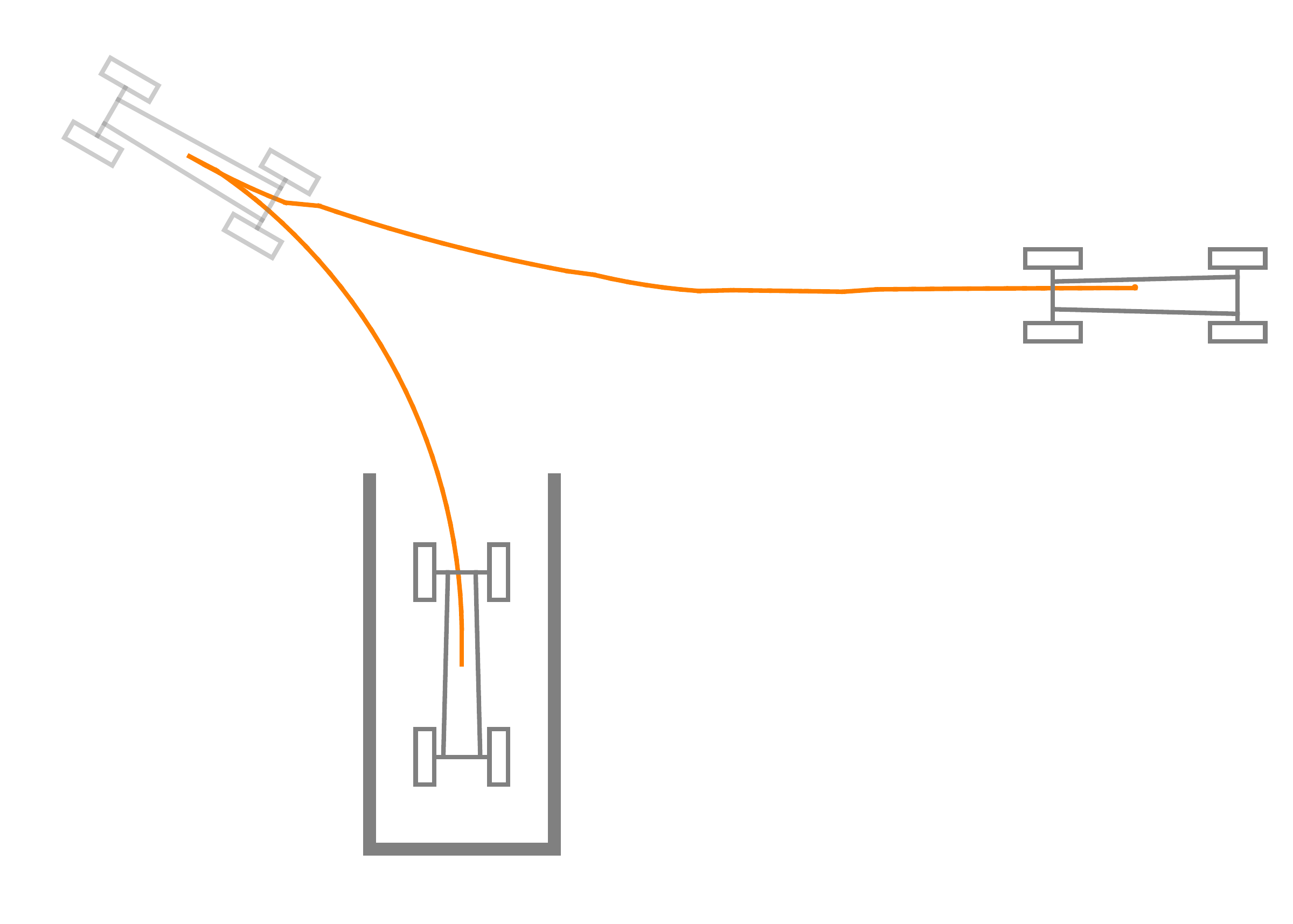}
	}\\
	\caption{Turning angles and facing angles of the car during parking. (a), (b) show the turning angles of the front wheel before and after smoothing procedure. (c), (d) show the facing angles of the car before and after smoothing procedure. (e), (f) show the corresponding path. Note that the path generated from target tree does not need to be smoothed. We will add some constrains in future to control the abrupt change of turning angle like in (b).}
	\label{fig:angle}
\end{figure}

One drawback of sample-based planning algorithms is that every segment of the path is in a random direction, which makes the path complex and causes redundant nodes. As mentioned before, There are already some smoothing algorithms for it. Both of these smoothing algorithms have a common drawback, they violently distort the path into a smooth curve, while not considering the kinematic constraint. As shown in Fig. \ref{fig:comp_smo}, the distortion can result in collision at some point. Our smoothing algorithm used the point\_pursuit algorithm to re-plan a path based on the path generated by RRT. As mentioned in part III, paths smoothed by our algorithm will be feasible. The path before and after smoothing is presented in Fig. \ref{fig:comp_smo}. We can see the path is much smoother after the procedure. From the car's moving perspective, in Fig. \ref{fig:angle}, when it is maneuvering in the pre-smoothed path, the turning angle of the front wheel will shake between $-30^{\circ}$ and $30^{\circ}$, the facing angle of the car will also shake. But these two angles will be far more stable and human-like in the smoothed path.

\subsection{Complex Parking Scenarios}
\begin{figure}[!h]
	\centering
	\subfloat[]{
		\includegraphics[width=4cm]{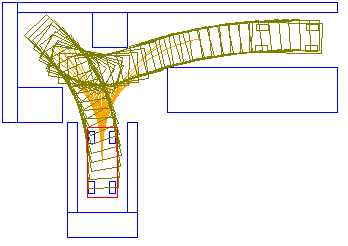}
	}
	\subfloat[]{
		\includegraphics[width=4cm]{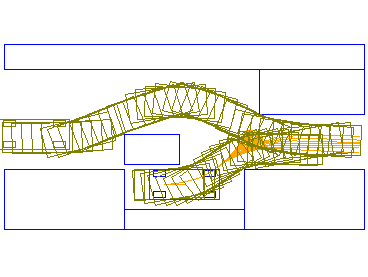}
	}\\
	
	\caption{This figure shows some paths planed by our algorithm in super narrow parking scenarios. The red rectangle is the destination.}
	\label{fig:narrow}
\end{figure}

We have tested our algorithm in various complex parking scenarios. Including narrow slot of highly limited space. As shown in Fig. \ref{fig:narrow}, our algorithm works very well in such an environment.

\section{Conclusion and future works}
This paper introduced imagination into RRT algorithm to expand the target point to a model-based target tree, which makes the algorithm quicker and the generated trajectory smoother. We also proposed a smoothing algorithm considering vehicle kinematic constraints to generate feasible trajectory.

\addtolength{\textheight}{-8cm}   
Our method still has some drawbacks, for example our car has to drive with a fixed speed. In the future we will use deep reinforcement learning to generalize the vehicle speed to a continuous value and explore an unknown space to locate an available parking lot. We will also test our algorithm in the real world.

\bibliographystyle{IEEEtran}
\bibliography{ref}

\end{document}